\documentclass{article}
\usepackage{preprint,times}

\usepackage{amsmath,amsfonts,bm}

\def\eqref#1{equation~\ref{#1}}

\def\1{\bm{1}}

\DeclareMathAlphabet{\mathsfit}{\encodingdefault}{\sfdefault}{m}{sl}
\SetMathAlphabet{\mathsfit}{bold}{\encodingdefault}{\sfdefault}{bx}{n}

\usepackage{hyperref}
\usepackage{url}
\usepackage{booktabs}
\usepackage{xcolor}
\usepackage{graphicx}
\usepackage{cleveref}
\usepackage{amssymb}
\usepackage{pifont}
\usepackage{array}
\usepackage{longtable}
\usepackage{siunitx}

\title{Samples, Sources, Space: Decomposing Data Scale in Spatially Structured Representation Learning of Human Brain Microarchitecture}

\author{
Christian Schiffer$^{1,2}$,
Mathis Bode$^{3}$,
Thomas Lippert$^{3,4}$,
Katrin Amunts$^{1,5}$,
Timo Dickscheid$^{1,2,6}$
\\[0.5em]
$^{1}$Institute of Neuroscience and Medicine (INM-1), Research Centre Jülich, Jülich, Germany
\\
$^{2}$Helmholtz AI, Research Centre Jülich, Jülich, Germany
\\
$^{3}$Jülich Supercomputing Centre, Research Centre Jülich, Jülich, Germany
\\
$^{4}$Frankfurt Institute for Advanced Studies, Goethe University Frankfurt, Frankfurt am Main, Germany
\\
$^{5}$Cécile \& Oscar Vogt Institute for Brain Research, University Hospital Düsseldorf, Düsseldorf, Germany
\\
$^{6}$Computer Vision, Institute for Computational Visualistics, University of Koblenz, Koblenz, Germany
\\[0.5em]
Correspondence: \texttt{c.schiffer@fz-juelich.de}
}

\hypersetup{pdftitle={Samples, Sources, Space: Decomposing Data Scale in Spatially Structured Representation Learning of Human Brain Microarchitecture},
  pdfauthor={Christian Schiffer, Mathis Bode, Thomas Lippert, Katrin Amunts, Timo Dickscheid}}

\begin{document}
\maketitle

\begin{abstract}
	Scaling studies typically represent training data by a single count of samples.
	For hierarchically and spatially structured data, however, the same number of samples can be drawn from few or many sources and distributed differently across the underlying domain.
	We therefore study data scaling as an allocation problem, separating unique sample count, source diversity, and spatial coverage.
	We study this decomposition in microscopic whole-brain histology, where a source is an individual brain, and a sample is an image patch at a specific spatial location.
	Across 93 controlled pretraining runs of a contrastive model that uses spatial proximity for supervision, we vary data allocation, compute, and model capacity over 11.6 million spatially anchored image patches from 21 human brains.
	Performance improves with more unique samples, broader spatial coverage, additional compute, and larger model capacity.
	At fixed sample count, distributing samples across one to 18 subjects produces no detectable improvement, even though representations generalize substantially better to subjects encountered during pretraining.
	Inter-subject variation therefore strongly affects generalization, but additional subjects provide no benefit when a fixed sample budget is distributed across more sources.
	These results establish sample count, source diversity, and spatial coverage as distinct axes of data scaling in spatially structured representation learning.
\end{abstract}

\section{Introduction}

Empirical scaling studies characterize how learning changes with model capacity, training compute, and dataset size, typically representing data scale by a single count of training examples $N$ \citep{Kaplan2020,Hoffmann2022,Henighan2020}.
In many scientific datasets, however, samples are \emph{hierarchically} and \emph{spatially} organized, so the same $N$ can correspond to substantially different data compositions.
Hierarchical organization determines how samples are distributed across sources: increasing $N$ may mean sampling existing sources more densely or acquiring additional sources.
Spatial organization determines how samples are distributed over the underlying domain: at fixed sample and source counts, samples may densely cover a restricted region or sparsely cover a broader one.
Source diversity, sampling density, and domain coverage are therefore distinct properties of a dataset.

This raises a basic question:
\emph{What does it mean to add data?}
We characterize data composition by the number of distinct samples $N_{\mathrm{unique}}$, the number of contributing sources $B$, and domain coverage $\rho$.
A \emph{unique sample} is one distinct sampled position, counted once irrespective of how often it is presented to the optimizer.
The resulting allocation problem is to determine what is gained by additional samples, additional sources, or broader domain coverage under controlled training compute and model capacity.

Variants of this problem arise across fields.
In pathology, patch count competes with diversity across slides~\citep{Bosch2025}, while ecological sampling allocates fixed effort between replication within locations and coverage of an environmental gradient~\citep{Schweiger2026}.
Analogous hierarchical and spatial structure also occurs in remote sensing~\citep{Wu2025} and spatial transcriptomics~\citep{Xu2023}.
Across these settings, the methodological question is whether different ways of increasing data are interchangeable or have distinct effects on learning.
Answering it requires the corresponding factors to be varied independently.

\begin{figure}[t]
	\centering
	\includegraphics[width=\textwidth]{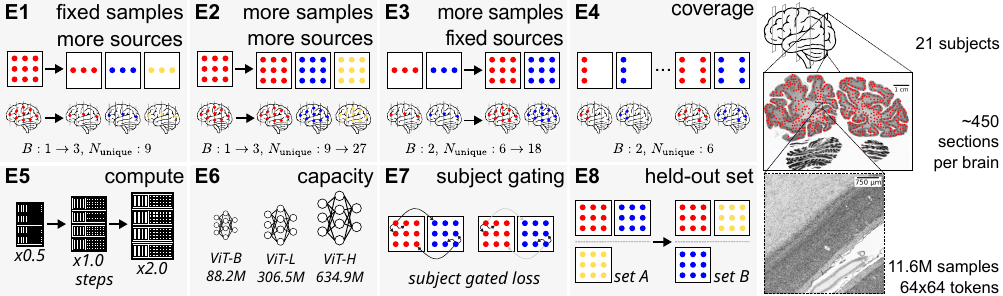}
	\caption{\textbf{Experimental decomposition of data scale and its instantiation in whole-brain histology.}
		\textbf{E1--E4} vary the composition of the training data by redistributing a fixed number of samples across more sources, increasing samples together with sources, increasing samples within fixed sources, or changing spatial coverage at fixed sample and source counts.
		The lower pictograms show the corresponding manipulations for brain histology, where sources are individual subjects and samples are spatially anchored image patches.
		\textbf{E5--E8} provide complementary controls over training compute, model capacity, cross-subject contributions to the SpatialNCE objective, and the choice of held-out evaluation subjects.
		See \cref{tab:axes} for details.
		The data hierarchy on the right maps the abstract units to the imaging pipeline from whole brains, to histological section, and image patches.
	}
	\label{fig:schematic}
\end{figure}

In this study, we investigate this problem in microscopic whole-brain histology, a particularly informative setting for controlled variation of sample count, source diversity, and spatial coverage.
In this setting, whole brains are sectioned, stained, digitized, and represented by image patches sampled from the resulting sections \citep{Amunts2013,Amunts2020,Schiffer2025}.
A source is an individual subject, a sample is an image patch, and the domain is anatomical space (\cref{fig:schematic}).
Each brain provides millions of samples, whereas adding another source requires processing an entire postmortem brain.
Its explicit spatial organization provides a learning signal for representation learning without manual labels~\citep{Schiffer2025}.
We consider 11.6 million sample locations from 21 human brains across 93 controlled pretraining runs spanning ViT-B to ViT-H, varying unique sample count, source count, spatial coverage, training compute, and model capacity.
Representations are evaluated by linear probing for classification of cytoarchitectonic areas~\citep{Amunts2020}.

Our experiments reveal a distinction between variation that affects generalization and variation that is valuable to acquire under a fixed sampling budget.
Representations generalize substantially better to subjects encountered during pretraining, showing that capturing inter-subject variation is critical for the learned representation.
However, redistributing a fixed number of samples across more subjects produces no detectable improvement across the tested model capacities.
The value of additional sources therefore depends on how they trade off against sampling density, rather than following directly from the magnitude of source-level variation.

More broadly, increasing dataset size can correspond to qualitatively different changes in data composition.
Additional samples, additional sources, and broader domain coverage need not contribute equally to representation quality.
We introduce a controlled experimental decomposition that makes these contributions separately measurable under matched compute and model capacity.
This turns data scaling in hierarchically and spatially structured datasets into an explicit allocation problem and provides a framework for identifying which form of additional data is most valuable.

\section{Related work}

\paragraph{Scaling laws and the data axis.}
Empirical scaling laws relate performance to model size, dataset size, and compute~\citep{Kaplan2020,Henighan2020,Hoffmann2022}.
Data scale is typically represented by example or token count, without distinguishing additional sources from within-source samples.
Data-curation methods show that composition matters within a fixed count through deduplication, pruning, and sample selection~\citep{Lee2022,Sorscher2023,Tirumala2023}, but do not decompose scale into source diversity, within-source density, and spatial coverage.
Closest to our setting, \citet{Muennighoff2023} study uniqueness versus repetition when data are constrained, whereas we ask how a fixed budget of unique samples should be allocated across sources and space.

\paragraph{Hierarchical data and sampling allocation.}
Many scientific datasets contain observations nested within larger acquisition units, creating a trade-off between denser sampling within sources and broader sampling across sources.
In computational pathology, tiles are nested within slides, specimens, and patients, and foundation models are pretrained on large numbers of both tiles and sources~\citep{Chen2024,Vorontsov2024,Lu2024}.
Cross-model evidence suggests that neither slide nor tile count alone explains downstream performance~\citep{Campanella2025}, but these comparisons jointly vary architectures, objectives, datasets, and compute and cannot isolate source diversity from within-source density.
Related allocation problems arise in ecology, where fixed sampling effort is divided between replication within locations and coverage across environmental gradients~\citep{Schweiger2026}, and in geospatial learning, where observations are spatially clustered and acquisition must be distributed across locations~\citep{Betti2026}.
Sample dependence is also recognized through subject-level splitting, spatial generalization benchmarks, and avoidance of pseudoreplication~\citep{Yagis2021,Butsko2025,Hurlbert1984}.
These works establish that sampling structure matters, but not how to allocate a fixed pretraining budget across samples, sources, and spatial coverage.

\paragraph{Spatially supervised representation learning.}
Tile2Vec~\citep{Jean2018}, SGCL~\citep{Lin2023}, and related histological approaches~\citep{Oberstrass2024c} define positive relationships from spatial proximity.
SpatialNCE, introduced with CytoNet for brain histology~\citep{Schiffer2025} and used here, instead weights pairwise relationships continuously by spatial distance.
The spatial composition of the training set therefore determines the supervisory relationships available to the objective, motivating our treatment of spatial coverage as an explicit experimental factor.

\section{Study design}

\subsection{Data, task, and generalization levels}

The dataset comprises whole-brain histological images from 21 postmortem human subjects~\citep{Amunts2020}, sectioned at $20\mu\mathrm{m}$, stained for neuronal cell bodies~\citep{Merker1983}, and digitized at $1\mu\mathrm{m}$ per pixel (see \cref{app:tissue}).
Approximately every fifteenth physical section is used, yielding about 450 sections per brain across coronal and horizontal cutting planes, with the plane recorded per subject and used to stratify subject subsets.
Candidate sample locations were generated independently within each section from gray- and white-matter masks~\citep{Schiffer2025}, using greedy sampling based on repeated Euclidean distance transforms (see \cref{app:sampling}).
After quality control, this yielded \num{11644201} locations, approximately \num{554000} per subject.
Each location is mapped to the Colin27 reference space~\citep{Holmes1998}, so Euclidean distances are defined between samples from different subjects~\citep{Schiffer2025}, and centers a $2048\times2048$-pixel input patch read at $2\mu\mathrm{m}$ per pixel (approximately $4\times4\mathrm{mm}$) during pretraining.

Three subjects were excluded from every controlled pretraining run and from probe training, providing the primary evaluation subjects for separating pretraining from probe exposure.
The remaining 18 subjects form the pretraining pool, of which 17 carry brain area annotations.
The downstream task is classification of 191 cytoarchitectonically defined areas using expert annotations from the Julich Brain Atlas~\citep{Amunts2020}.
We train a linear classifier on frozen representations, compute macro-F1 over the classes present in each evaluation subject, and then average across subjects.
The class set is fixed before evaluation, and probe optimization is identical for every encoder.
Probe-training sets contain at most \num{1000} samples per class, selected without replacement and round-robin across subjects.
No physical section is shared between probe training and evaluation.
Full probe-training and evaluation details are provided in \cref{app:probe}.
Generalization level is defined per model--evaluation-subject pair, since a subject may be in the pretraining set of one model and excluded from that of another.
An \emph{unseen} subject is absent from both pretraining and probe training, a \emph{transfer} subject is present in pretraining and absent from probe training, and a \emph{held-out-section} subject contributes other sections to both while the evaluated sections remain held out.

\subsection{Objective and models}

All encoders are pretrained with SpatialNCE \citep{Schiffer2025}, which encourages patches from nearby anatomical locations to have more similar representations than patches from distant ones.
Let $x_i$ denote an image patch, $p_i\in\mathbb{R}^3$ its 3D reference location, and $z_i$ the normalized output of encoder and projection head.
For a batch, the loss for sample $i$ and the continuous target weight between two samples are given by
\begin{equation}
	\mathcal{L}_{i} = - \frac{1}{\sum_{j \neq i} \omega_{ij}} \sum_{j \neq i} \omega_{ij} \log \frac{ \exp\left( z_i^\top z_j / \tau \right) }{\sum_{k \neq i} \exp\left( z_i^\top z_k / \tau \right)},
	\qquad
	\omega_{ij} = \exp\left(- \frac{\lVert p_i-p_j\rVert_2^2}{2\sigma^2}\right).
\end{equation}
Following \citet{Schiffer2025} we use $\sigma=10\,\mathrm{mm}$, $\tau=0.1$, and compute the objective over all non-identical pairs in a global batch of \num{2048}, gathered across all accelerators.
Coordinates determine the supervisory relationship and are never given to the encoder, which must infer spatially consistent structure from image content alone.
The kernel contains no subject indicator, so anatomically corresponding samples receive high weight regardless of subject identity.
The normalizing term $\sum_{j\neq i}\omega_{ij}$ is the total positive kernel weight associated with anchor $i$ in a batch, and we call its average across anchors and batches the \emph{positive weight} $M$.
$M$ depends only on the data composition and the batching procedure, so it can be computed before training.
Dataset composition therefore changes the effective objective: concentrating samples spatially increases $M$, while changing subject count alters the balance of within- and cross-subject pairs in the kernel.
To isolate the role of cross-subject correspondences, one experiment (\textbf{E7}) sets all cross-subject weights to zero.
We refer to this modification as \emph{subject gating}.

We use encoders ViT-B, ViT-L, and ViT-H following the ViT-5 modernization of plain Vision Transformers \citep{Dosovitskiy2020,Wang2026}.
Each input is tokenized into a $64\times64$ grid of \num{4096} tokens, together with a class token and four register tokens \citep{Darcet2023}.
All models train with AdamW~\citep{Loshchilov2018} at a constant learning rate \num{1e-4} on 64 to 256 NVIDIA GH200 accelerators (see \cref{app:env,app:training} for details).
The 93 runs consumed \num{316000} accelerator-hours over $7.32$ million optimization steps and \num{94200} training exaFLOPs.
Including the 124 linear probes, the study totals \num{349000} accelerator-hours.
Across comparisons, the architecture family, optimizer, batch size, and evaluation protocol are held fixed, so experiments isolate changes in data composition, compute, or model capacity.
Architecture dimensions, augmentations, optimizer settings, and FLOP accounting are given in Appendices~\ref{app:arch} and \ref{app:compute}.

\subsection{Experiments}\label{sec:experimental_design}

We organize the experiments around five factors, describing representation quality as $S=f(P,C,N_{\mathrm{unique}},B,\rho)$.
$S$ is macro-F1 on subjects excluded from both pretraining and probe training.
$P$ denotes model capacity, $C$ compute measured in optimization steps, $N_{\mathrm{unique}}$ the number of distinct sample locations, $B$ the number of contributing sources, and $\rho$ the spatial coverage condition.
Sources are individual subjects here, so $B$ is a subject count throughout the results.
The data axes are not fully independent: at fixed $N_{\mathrm{unique}}$, increasing $B$ necessarily reduces the mean number of locations per source $N_{unique}/B$.
\Cref{tab:axes} summarizes the 93 runs.
Unless compute is the manipulated factor, runs use a default budget of \num{73200} steps, or \num{149913600} sample presentations.
Presentations per unique location therefore vary inversely with $N_{\mathrm{unique}}$, from roughly \num{3400} in the smallest single-subject condition to roughly 15 in the largest, and equal 333 throughout \textbf{E1}.
The unique-count axis is therefore an axis of \emph{uniqueness at a fixed presentation budget}, which differs from the data term in compute-optimal scaling laws \citep{Hoffmann2022,Muennighoff2023}.
Subject subsets were chosen before any downstream result existed, from subject metadata with a fixed seed, stratified by cutting plane.
For each level of $B$, we construct as many disjoint groups as the pool permits ($\lfloor 18/B\rfloor$), capped at two groups for the fixed-sample experiment \textbf{E1} and four for the more-sources-more-samples experiment \textbf{E2}.
Each configuration is trained with a single optimization seed, so replication arises from the independently composed subject groups.
Their spread therefore captures sensitivity to \emph{which} subjects are sampled.

\begin{table}[t]
	\caption{
		\textbf{Overview of the eight experiments and 93 pretraining runs.}
		Run counts are incremental, with reused configurations counted only once.
		Unless stated otherwise, runs use ViT-B and the default presentation budget.
		In this study, a source corresponds to a subject, and a section is an ordered spatial subdivision of that source along the anatomical domain.
	}
	\label{tab:axes}
	\centering
	\begin{tabular}{@{}lp{0.16\textwidth}p{0.35\textwidth}p{0.25\textwidth}r@{}}
		\toprule
		            & \textbf{Experiment}                                    & \textbf{Manipulation}                                          & \textbf{Levels} & \textbf{Runs} \\
		\midrule
		\textbf{E1} & Fixed samples, more sources                            & redistribute $N_{\mathrm{unique}}=\num{450000}$ across sources
		            & $B\in\{1,2,4,8,18\}$                                   & 9                                                                                                \\\addlinespace[3pt]
		\textbf{E2} & More samples, more sources                             & retain selected sections as $B$ increases
		            & $B\in\{1,2,4,8,18\}\times\{$all, every 10th$\}$        & 30                                                                                               \\\addlinespace[3pt]
		\textbf{E3} & More samples, fixed sources                            & vary retained-section fraction at fixed $B$
		            & $B\in\{1,9,18\}\times\{$all, every 3rd, every 10th$\}$ & 5                                                                                                \\\addlinespace[3pt]
		\textbf{E4} & Coverage                                               & arrange fixed $N_{\mathrm{unique}}$ into $k$ section blocks
		            & $B\in\{1,4,13\}$, $k\in\{$every 10th, 16, 4, 1$\}$     & 19                                                                                               \\\addlinespace[3pt]
		\textbf{E5} & Compute                                                & vary optimization steps
		            & $0.5\times$, $1\times$, $2\times$                      & 7                                                                                                \\\addlinespace[3pt]
		\textbf{E6} & Capacity                                               & vary encoder size
		            & ViT-B, ViT-L, ViT-H                                    & 14                                                                                               \\\addlinespace[3pt]
		\textbf{E7} & Subject gating                                         & zero cross-subject kernel weights
		            & $B\in\{2,4,8,18\}$                                     & 5                                                                                                \\\addlinespace[3pt]
		\textbf{E8} & Held-out sources                                       & vary excluded source triple
		            & three held-out assignments at $B=18$                   & 4                                                                                                \\
		\bottomrule
	\end{tabular}
\end{table}

Experiments \textbf{E1}--\textbf{E4} define data compositions.
\textbf{E5}--\textbf{E8} are crossed with a subset of them, so a change in compute, capacity, gating, or held-out assignment is always measured against an otherwise identical composition.
The corresponding detection thresholds are defined in \cref{sec:stats}.
\Cref{app:inventory_table} lists every run.
\emph{Fixed samples, more sources} (\textbf{E1}) sets $N_{\mathrm{unique}}=\num{450000}$ and varies $B\in\{1,2,4,8,18\}$, selecting sections at equidistant intervals within each subject so that broad anatomical coverage is preserved as within-subject sampling density decreases.
\emph{Coverage} (\textbf{E4}) holds $N_{\mathrm{unique}}$ fixed and varies how those locations are arranged along each subject's section series.
Every subject contributes $k$ disjoint contiguous section blocks, grown outward from evenly spaced anchors until they reach the target sample budget, so larger $k$ yields more and shorter blocks and therefore broader coverage.
At $k=1$, a single localized block is placed at an anterior, central, or posterior anchor, and these three positions are analyzed separately.
The intermediate levels matter because anatomical coverage and positive weight $M$ change at different rates as $k$ decreases.
16 blocks preserve nearly the full anatomical extent and leave $M$ almost unchanged, whereas four blocks introduce larger gaps and raise $M$ by half.
The primary contrast is 16 blocks against every-10th-section sampling at $B=13$, which matches both $N_{\mathrm{unique}}$ and $M$ and isolates the effect of introducing spatial gaps.
\emph{Subject gating} (\textbf{E7}) leaves within-subject distances unchanged and sets cross-subject weights to zero, so gated and standard objectives coincide at $B=1$ and diverge as $B$ grows.
\emph{Held-out sources} (\textbf{E8}) uses two additional disjoint triples of held-out subjects.
Existing eligible models are re-evaluated on these triples.
Four additional models are trained with the alternative triples excluded from pretraining.

\subsection{Statistical analysis}\label{sec:stats}

Each of the three primary evaluation subjects contributes one score to every configuration.
Let $S_{cd}$ denote the macro-F1 score of configuration $c$ on evaluation subject $d$.
We fit a linear mixed-effects model to logit-transformed scores across the experimental grid:
\begin{equation}
	\begin{aligned}
		\operatorname{logit}(S_{cd}) & =
		\beta_0
		+\beta_N\log_2 N_{\mathrm{unique},c}
		+\beta_B\log_2 B_c
		+\beta_C\log_2 C_c
		+\beta_M\log_2 M_c               \\
		                             &
		+\beta_G G_c
		+\gamma_P(P_c)
		+\gamma_{\rho}(\rho_c)
		+u_{\mathrm{subset}(c)}+v_d+e_{cd}.
	\end{aligned}
\end{equation}
Because macro-F1 is bounded to $[0,1]$, we model its logit transformation, so coefficients are expressed on the logit scale.
Continuous predictors enter as base-two logarithms, so their coefficients quantify the change in logit macro-F1 associated with a doubling of the corresponding quantity.
Model capacity $P_c$ and coverage condition $\rho_c$ enter as categorical fixed effects through $\gamma_P$ and $\gamma_{\rho}$, respectively, and $G_c$ is an indicator for subject gating.
The random intercept $u_{\mathrm{subset}(c)}$ accounts for differences among pretraining-subject subsets, while $v_d$ accounts for systematic differences among evaluation subjects.
A second model is restricted to the fixed-sample experiment and adds a capacity-by-subject-count interaction, $\gamma_{BP}(P_c)\log_2 B_c$, testing whether the effect of increasing $B$ differs by model capacity.
The full-grid fit uses 96 configurations and 288 subject-level scores, comprising 93 final pretraining runs and three retained intermediate checkpoints used as shorter-compute conditions in experiment \textbf{E5}.
The fixed-sample fit uses 18 configurations and 54 scores.
Both models are fitted by restricted maximum likelihood.
We report 95\% confidence intervals throughout, using profile-likelihood intervals for the scaling coefficients and capacity interaction and Wald intervals elsewhere.\footnote{Estimates are reported as $\text{coefficient}=\text{estimate}\,[\text{lower},\,\text{upper}]$, on the logit scale per doubling unless stated otherwise.}

To distinguish unresolved differences from effects that the design was capable of detecting, we define minimum detectable effects (MDEs) from the fitted variance components at 80\% power.
The MDE is $0.226$ logit for contrasts across pretraining-subject subsets and $0.166$ logit for contrasts within the same subset.
Differences below the applicable MDE are reported as \emph{unresolved}.
These thresholds apply to pairwise configuration contrasts and not to regression coefficients, which are evaluated using their confidence intervals.
The MDE procedure and thresholds were fixed before interpreting the experimental contrasts, and primary and exploratory analyses are specified in \cref{app:prespec}.
\Cref{app:stats_sensitivity} gives the full MDE derivation, and \cref{app:vif} reports collinearity diagnostics for the fixed-effect design.

\section{Results}

\subsection{Source inclusion in pretraining dominates performance}\label{sec:r1}

\begin{figure}[t]
	\centering
	\includegraphics[width=\textwidth]{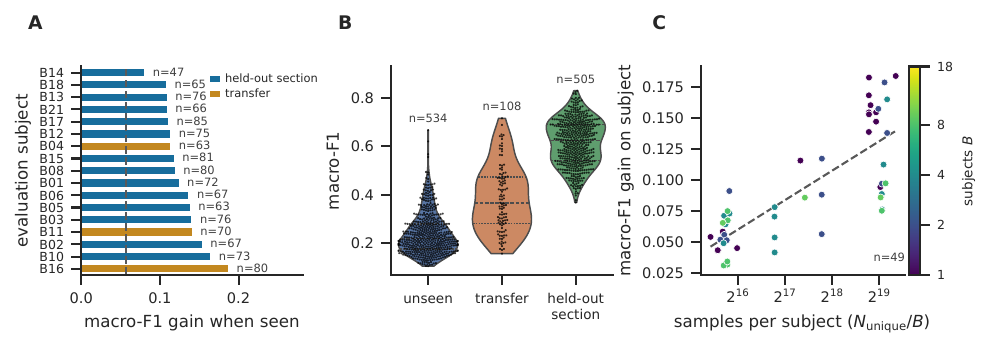}
	\caption{\textbf{Subject inclusion in pretraining dominates performance.}
		\textbf{A}: Macro-F1 score gain when a subject was included in the pretraining pool of a model vs. when it was not; averaged over $n$ models (gray line: MDE $0.057$), at identical evaluation samples, labels, and class denominator.
		\textbf{B}: Score distribution by generalization level: unseen subjects are absent from pretraining and probe training, transfer subjects only from probe training, and held-out-section subjects contribute other sections to both.
		\textbf{C}: Familiarity gain against $N_{\mathrm{unique}}/B$, the mean number of locations a model saw per subject.
		Each point is one model from \textbf{E1}, \textbf{E2}, \textbf{E3}, \textbf{E6}, giving its subject-centered macro-F1 gain on pretraining subjects over other subjects.
	}\label{fig:generalization}
\end{figure}

Whether an evaluation subject was encountered during pretraining had the largest observed effect on performance among the factors examined here (\cref{fig:generalization}).
We compare configurations pretrained with a given evaluation subject against configurations pretrained without that subject, while holding evaluation samples, labels, probe protocol, and class denominator fixed.
Across the 14 held-out-section subjects, macro-F1 was higher by $0.124$ on average when the subject had been seen during pretraining, with subject-level differences from $0.081$ to $0.165$.
Across the three transfer subjects, the mean advantage was $0.148$, with differences from $0.114$ to $0.188$.
Configurations with and without a given subject necessarily use different pretraining-subject subsets, so the across-subset MDE of $0.226$ logit applies.
All 17 subject-level contrasts were positive and exceeded the applicable MDE on the logit scale, and 14 of 17 exceeded twice that threshold.

The number of familiar evaluation subjects increases almost deterministically with $B$, so a cross-model regression cannot identify whether the gap itself changes with subject count.
We therefore examine it against $N_{\mathrm{unique}}/B$, the mean number of unique locations seen per pretraining subject.
Across 49 models the familiarity advantage grows by $0.024$ macro-F1 per doubling of $N_{\mathrm{unique}}/B$, and in a joint model that coefficient remains $0.022\,[0.017,0.027]$ while the coefficient for $\log_2 B$ is $-0.012\,[-0.019,-0.005]$.
The advantage is therefore associated more strongly with per-subject sampling density than with subject count, which sharpens the question for the next experiment: whether more subjects improve generalization when the total sample budget is held fixed.

\subsection{More sources do not help at a fixed sample budget}\label{sec:r2}

\begin{figure}[t]
	\centering
	\includegraphics[width=\textwidth]{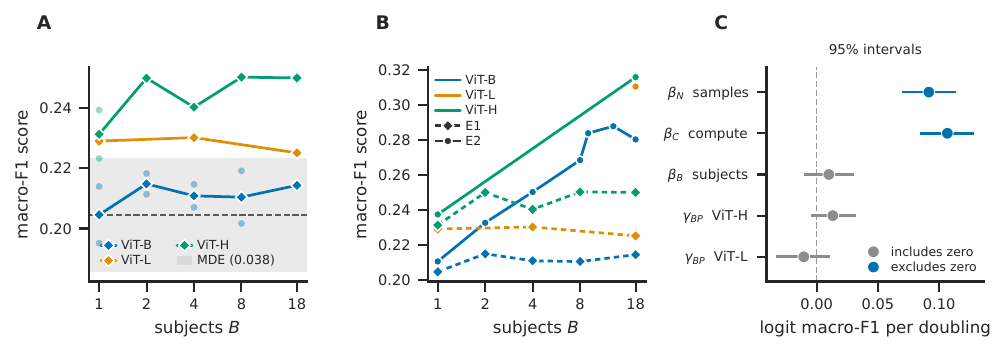}
	\caption{\textbf{Scaling subject count $B$ without scaling $N_\mathrm{unique}$ does not improve representations.}
		\textbf{A}: Macro-F1 against subject count $B$ at fixed $N_{\mathrm{unique}}\!\approx\!\num{450000}$ (\textbf{E1} in \cref{tab:axes}), with solid lines giving the mean per capacity $P$ (color) and the MDE band anchored at the $B{=}1$ mean (dashed line).
		The curve is flat at every capacity.
		\textbf{B}: The same axis overlaid with \textbf{E2}, where $N_{\mathrm{unique}}$ grows linearly with $B$.
		\textbf{C}: 95\% intervals of coefficients per doubling on the logit scale.
		Intervals for $\beta_B$ and the capacity interaction $\gamma_{BP}$ include zero, while performance improves with sample count $\beta_{N}$ and compute $\beta_{C}$.
	}\label{fig:equal_sample}
\end{figure}

Distributing a fixed sample budget across more subjects produced no detectable improvement in representation quality (\cref{fig:equal_sample}).
For ViT-B in the fixed-sample analysis, the estimated subject-count effect was $\beta_B=0.0014\,[-0.0253,0.0282]$ logit macro-F1 per doubling of $B$.
Across the full range from $B=1$ to $B=18$ ($\log_2 18=4.17$ doublings), this interval bounds the fitted change to $[-0.106,0.118]$ logit, or approximately $[-0.017,0.020]$ macro-F1 at the reference score.
The observed macro-F1 differences between $B=1$ and $B=18$ were $0.015$ for ViT-B, $-0.002$ for ViT-L, and $0.017$ for ViT-H, all below the across-subset MDE that applies to these contrasts.
The subject-count effect also did not increase detectably with model capacity: relative to ViT-B, the capacity-by-subject-count interaction was $\gamma_{BP}=0.0132\,[-0.0046,0.0321]$ for ViT-H and $-0.0105\,[-0.0328,0.0112]$ for ViT-L.
The full-grid model gives the same conclusion, with $\beta_B=0.0100\,[-0.0105,0.0306]$.
The sensitivity of \textbf{E1} to capacity-by-subject-count interactions is quantified by simulation in \cref{app:power}.

Two observations constrain the interpretation of this null result.
First, the detection threshold was fixed in advance, and the upper confidence limit for $\beta_B$ is only about one quarter of the estimated effect of doubling $N_{\mathrm{unique}}$ reported below.
Second, performance was still responsive to additional unique samples within the same subject pool: holding the same 18 subjects fixed while increasing $N_{\mathrm{unique}}$ from \num{450000} to $9.8$ million raised macro-F1 from $0.214$ to $0.300$, well above the applicable MDE.
The conclusion is therefore limited to the tested regime of $B=1$--$18$, approximately \num{450000} unique locations, SpatialNCE pretraining, and cytoarchitectonic targets.
The transferable result is the distinction itself: source count $B$ and coverage $\rho$ are separate experimental quantities, and neither follows from $N_{\mathrm{unique}}$.
Together with \cref{sec:r1}, inter-subject variability therefore strongly affects generalization, yet increasing subject diversity by reducing the number of samples per subject does not compensate for that variability.
The main conclusions were unchanged under alternative held-out-subject assignments (\cref{app:altholdout}).

\subsection{More samples, compute, and capacity improve representations}\label{sec:r3}

\begin{figure}[t!]
	\centering
	\includegraphics[width=\textwidth]{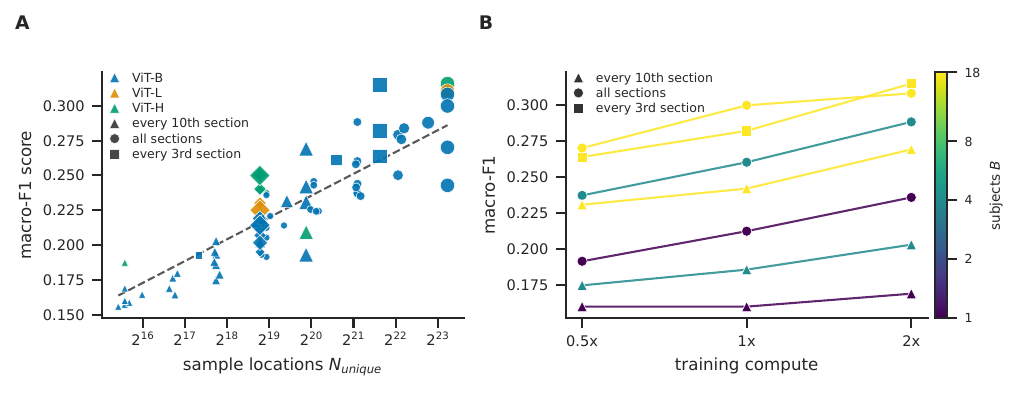}
	\caption{\textbf{Scaling samples $N_\mathrm{unique}$, compute $C$, and capacity $P$ improve representations.}
		\textbf{A}:~Macro-F1 against $N_{\mathrm{unique}}$ across the grid ($\beta_N = 0.09$ / doubling, dashed line), with marker color, style, and size giving capacity $P$, coverage $\rho$, and subject count $B$.
		\textbf{B}:~Macro-F1 against compute $C$, subject count $B$ (color), and coverage $\rho$ (marker style) across the 7 settings of \textbf{E5}.
	}
	\label{fig:scaling_axes}
\end{figure}

Increasing unique sample count, compute, and model capacity each improved representation quality (\cref{fig:scaling_axes}).
Doubling the number of unique sample locations increased logit macro-F1 by $\beta_N=0.0916\,[0.0695,0.1135]$, while doubling optimization steps increased it by $\beta_C=0.1067\,[0.0847,0.1286]$.
Because the presentation budget is fixed, $\beta_N$ captures the joint effect of doubling uniqueness while halving repetition, whereas the fixed-sample estimate of $\beta_B$ varies subject count at constant repetition.
All seven configurations evaluated at both the default and doubled compute budgets improved, by $0.008$ to $0.034$ macro-F1, and the second doubling gave roughly $43\%$ of the gain of the first.

Capacity point estimates were monotone.
At $B=18$, macro-F1 increased from $0.300$ for ViT-B to $0.311$ for ViT-L and $0.316$ for ViT-H, none of which reaches the within-subset MDE that applies to these matched-subset contrasts.
Evidence for a capacity effect comes instead from the categorical capacity terms pooled across the experimental grid, with $\gamma_P=0.127\,[0.094,0.159]$ for ViT-H and $0.081\,[0.031,0.130]$ for ViT-L relative to ViT-B.

In practice, adding a subject increases both source diversity and the number of available samples.
Under full sampling, ViT-B macro-F1 increased from $0.210$ across the four single-subject subsets to $0.300$ at $B=18$, and under every-10th-section sampling from $0.160$ to $0.242$.
Adding subjects is therefore beneficial when they also increase the number of available samples, while \cref{sec:r2} shows that subject count alone does not explain the gain.
For context, the strongest published reference encoder under the same protocol reached $0.251$ macro-F1 (Virchow2; \cref{app:reference}).

\subsection{Broad spatial coverage improves representations}\label{sec:r4}

\begin{figure}[t]
	\begin{center}
		\includegraphics[width=\textwidth]{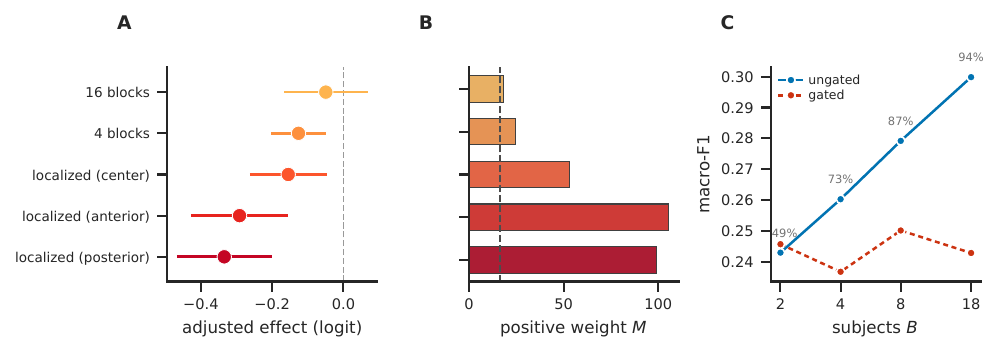}
	\end{center}
	\caption{\textbf{Coverage geometry, positive weight, and subject gating at fixed sample count.}
		\textbf{A}: Adjusted logit macro-F1 differences among the coverage conditions in \textbf{E4} relative to every-10th-section sampling, with 95\% confidence intervals, ordered by spatial concentration.
		\textbf{B}: Positive weight $M$ for the corresponding \textbf{E4} conditions, with every-10th-section sampling as the reference.
		\textbf{C}: Effect of subject gating in \textbf{E7}, which sets all cross-subject SpatialNCE weights to zero.
		Each gated configuration is paired with an ungated configuration using the same data composition.
		Annotations indicate the fraction of $M$ contributed by cross-subject pairs before gating.
		At $B=2$, where cross-subject pairs contribute $49\%$ of $M$, the gating contrast is unresolved.
		At $B=18$, where cross-subject pairs contribute $94\%$ of $M$, gating reduces macro-F1 by $0.057$.
	}
	\label{fig:s_coverage}
\end{figure}

Restricting a fixed sample budget to narrower anatomical coverage reduced representation quality at high subject count (\cref{fig:s_coverage}).
All $B=13$ coverage conditions use the same thirteen coronal subjects, so the within-subset MDE of approximately $0.028$ macro-F1 applies.
Relative to every-10th-section sampling at $B=13$, macro-F1 decreased by $0.038$ for four blocks, by $0.034$, $0.028$, and $0.021$ for the three localized single-block conditions, and by $0.017$ for 16 blocks.
The four-block condition and two of the three localized conditions meet or exceed the MDE, while the 16-block condition and the remaining localized condition do not.
Spatial concentration alone does not explain this ordering: the four-block condition spans $78\%$ of the full section range, yet shows the largest deficit, so the arrangement of covered and uncovered regions also contributes.
The primary contrast between 16 blocks and every-10th-section sampling matches both $N_{\mathrm{unique}}$ and positive weight $M$ and is unresolved, so introducing many small gaps into otherwise broad coverage has no detectable cost under this design.
The deficit also appears only at high subject count, since no coverage condition differs detectably from every-10th-section sampling at $B=1$ or $B=4$.

Most coverage conditions also change the positive weight $M$ and thus do not isolate anatomical coverage as a single factor.
At $B=13$, $M$ is $16.15$ for every-10th-section sampling and $17.94$ for 16 blocks, and ranges from $53.11$ to $105.58$ across the three localized conditions.
The localized conditions therefore combine reduced coverage with a substantial change in objective composition, so this experiment does not separate the two contributions.
Subject gating introduces an even larger change in positive weight.
Cross-subject pairs supply $94.2\%$ of the positive weight at $B=18$, so removing them reduces $M$ by roughly a factor of 17, and performance under gating was lower from $B=4$ onward.
At $B=18$, macro-F1 decreased from $0.300$ with the standard objective to $0.243$ with gating, a difference of $0.057$ that exceeds the within-subset MDE.
Across the full-grid model, doubling positive weight is associated with an increase of $\beta_M=0.0979\,[0.0597,0.1357]$ in logit macro-F1, consistent with much of that deficit.
Cross-subject anatomical correspondences therefore supply most of the positive mass available to SpatialNCE at high subject count and are important for maintaining performance in this regime.
The composition of $M$ across \textbf{E1}, \textbf{E4}, and \textbf{E7}, and the gating indicator after adjustment for $M$, are detailed in \cref{app:kernel,app:coverage}.

\section{Discussion}

A central implication of our results is that the magnitude of source-level variation and the value of acquiring additional sources are distinct quantities.
Inter-source variation strongly affects generalization, yet broader source sampling does not improve representations when it redistributes a fixed sample budget.
For hierarchical datasets, observing substantial variation between sources therefore provides insufficient evidence for how additional acquisition effort should be allocated.
The relevant quantity is the marginal value of samples from new versus already represented sources.

This distinction extends the usual notion of data scale.
A sample count alone does not specify whether dataset growth adds observations, sources, or coverage of the underlying domain, and our experiments show that these forms of scale can have different value within the same learning problem.
Making this decomposition explicit provides a way to reason about data acquisition in settings with repeated or spatially structured observations, where collecting another sample and collecting another source can have very different costs.
Our results establish this distinction in spatially supervised whole-brain histology and motivate treating data composition as an explicit dimension of empirical scaling studies.

\subsection*{Ethics statement}
The presented study requires no separate ethical approvals.
All usage in this work is covered by a vote of the ethics committee of the Medical Faculty of the Heinrich Heine University Düsseldorf (\#4863).
Postmortem brains were obtained through body donor programs of the anatomical institutes of the universities of Düsseldorf, Rostock, and Aachen, in accordance with legal and ethical regulations and guidelines.
All body donors have signed a declaration of agreement.

\subsection*{Reproducibility statement}
The analysis plan was pre-specified before the first pretraining run and fixed the unit of replication, mixed-effects specification, interval procedure, and minimum detectable effect.
Contrasts outside this plan are identified as exploratory.
\Cref{sec:experimental_design} and \cref{tab:axes} define all 93 pretraining runs, and \cref{app:inventory_table} lists them individually.
Code, configurations, and derived data were version-controlled with DataLad, providing provenance from pretraining manifests through the tables used to generate the reported results and figures.

\subsection*{Acknowledgments}

This project received funding from the European Union’s Horizon 2020 Research and Innovation Programme, grant agreement 101147319 (EBRAINS 2.0 Project), the Helmholtz Association portfolio theme “Supercomputing and Modeling for the Human Brain”, the Helmholtz Association’s Initiative and Networking Fund through the Helmholtz International BigBrain Analytics and Learning Laboratory (HIBALL) under the Helmholtz International Lab grant agreement InterLabs-0015, from HELMHOLTZ IMAGING, a platform of the Helmholtz Information \& Data Science Incubator [X-BRAIN, grant number: ZT-I-PF-4-061], and from the Deutsche Forschungsgemeinschaft (DFG, German Research Foundation) under the National Research Data Infrastructure – NFDI 46/1 – 501864659.
This project received access to the JUPITER supercomputer through the JUPITER Research and Early Access Program (JUREAP).
JUPITER is funded by the EuroHPC Joint Undertaking, the German Federal Ministry of Research, Technology and Space, and the Ministry of Culture and Science of the German state of North Rhine-Westphalia.

\subsection*{AI use statement}
Generative AI tools were used during the research workflow and manuscript preparation.
They assisted with discussion of experimental design and statistical analysis, refinement of the conceptual framing, manuscript drafting and editing, and generation and revision of analysis and figure code.
All AI-assisted text, code, analyses, and interpretations were reviewed by the authors.
All reported numerical results were verified against their generating data or analysis outputs, and AI-generated code was inspected and its outputs validated.
Model pretraining, evaluation, and the statistical fits reported in this work were produced by the authors' own computational pipelines.
The authors take responsibility for all content, analyses, figures, and conclusions in this manuscript.

\bibliography{references}

\begin{thebibliography}{36}
\providecommand{\natexlab}[1]{#1}
\providecommand{\url}[1]{\texttt{#1}}
\expandafter\ifx\csname urlstyle\endcsname\relax
  \providecommand{\doi}[1]{doi: #1}\else
  \providecommand{\doi}{doi: \begingroup \urlstyle{rm}\Url}\fi

\bibitem[Amunts et~al.(2013)Amunts, Lepage, Borgeat, Mohlberg, Dickscheid,
  Rousseau, Bludau, Bazin, Lewis, {Oros-Peusquens}, Shah, Lippert, Zilles, and
  Evans]{Amunts2013}
Katrin Amunts, Claude Lepage, Louis Borgeat, Hartmut Mohlberg, Timo Dickscheid,
  Marc-{\'E}tienne Rousseau, Sebastian Bludau, Pierre-Louis Bazin, Lindsay~B.
  Lewis, Ana-Maria {Oros-Peusquens}, Nadim~J. Shah, Thomas Lippert, Karl
  Zilles, and Alan~C. Evans.
\newblock {{BigBrain}}: {{An Ultrahigh-Resolution 3D Human Brain Model}}.
\newblock \emph{Science}, 340\penalty0 (6139):\penalty0 1472--1475, 2013.
\newblock ISSN 0036-8075, 1095-9203.
\newblock \doi{10.1126/science.1235381}.

\bibitem[Amunts et~al.(2020)Amunts, Mohlberg, Bludau, and Zilles]{Amunts2020}
Katrin Amunts, Hartmut Mohlberg, Sebastian Bludau, and Karl Zilles.
\newblock Julich-{{Brain}}: {{A 3D}} probabilistic atlas of the human brain's
  cytoarchitecture.
\newblock \emph{Science}, 369\penalty0 (6506):\penalty0 988, 2020.
\newblock \doi{10.1126/science.abb4588}.

\bibitem[Betti et~al.(2026)Betti, Sanni, Sogoyou, Agbagla, Molitor, Carleton,
  and Rolf]{Betti2026}
Livia Betti, Farooq Sanni, Gnouyaro~Z. Sogoyou, Togbe Agbagla, Cullen Molitor,
  Tamma Carleton, and Esther Rolf.
\newblock Mapping on a budget: {{Optimizing}} spatial data collection for
  {{ML}}.
\newblock In \emph{Proceedings of the {{AAAI Conference}} on {{Artificial
  Intelligence}}}, volume~40, pp.\  38233--38241, 2026.

\bibitem[Bosch et~al.(2025)Bosch, Wong, Paulikat, Zapukhlyak, Arora,
  {Aichm{\"u}ller-Ratnaparkhe}, Baumann, Karn, Kamble, Karnik, Khedkar, Chhut,
  Aswolinskiy, and Aichm{\"u}ller]{Bosch2025}
Christoph Bosch, John K.~L. Wong, Martin Paulikat, Myroslav Zapukhlyak, Bharti
  Arora, Manasi {Aichm{\"u}ller-Ratnaparkhe}, Jens Baumann, Shivani Karn,
  Rutuja Kamble, Swapnil Karnik, Bhushan Khedkar, Serey~Vathana Chhut, Witali
  Aswolinskiy, and Christian Aichm{\"u}ller.
\newblock Diversity {{Over Scale}}: {{Whole-Slide Image Variety Enables
  H}}\&{{E Foundation Model Training}} with {{Fewer Patches}}, November 2025.

\bibitem[Butsko et~al.(2025)Butsko, Tricht, Tseng, Milli, Rolnick, Cartuyvels,
  Reshef, Szantoi, and Kerner]{Butsko2025}
Christina Butsko, Kristof~Van Tricht, Gabriel Tseng, Giorgia Milli, David
  Rolnick, Ruben Cartuyvels, Inbal~Becker Reshef, Zoltan Szantoi, and Hannah
  Kerner.
\newblock Deploying {{Geospatial Foundation Models}} in the {{Real World}}:
  {{Lessons}} from {{WorldCereal}}, July 2025.

\bibitem[Campanella et~al.(2025)Campanella, Chen, Singh, Verma, Muehlstedt,
  Zeng, Stock, Croken, Veremis, Elmas, Shujski, Neittaanm{\"a}ki, Huang, Kwan,
  Houldsworth, Schoenfeld, and Vanderbilt]{Campanella2025}
Gabriele Campanella, Shengjia Chen, Manbir Singh, Ruchika Verma, Silke
  Muehlstedt, Jennifer Zeng, Aryeh Stock, Matt Croken, Brandon Veremis,
  Abdulkadir Elmas, Ivan Shujski, Noora Neittaanm{\"a}ki, Kuan-lin Huang, Ricky
  Kwan, Jane Houldsworth, Adam~J. Schoenfeld, and Chad Vanderbilt.
\newblock A clinical benchmark of public self-supervised pathology foundation
  models.
\newblock \emph{Nature Communications}, 16\penalty0 (1):\penalty0 3640, April
  2025.
\newblock ISSN 2041-1723.
\newblock \doi{10.1038/s41467-025-58796-1}.

\bibitem[Chen et~al.(2024)Chen, Ding, Lu, Williamson, Jaume, Song, Chen, Zhang,
  Shao, Shaban, Williams, Oldenburg, Weishaupt, Wang, Vaidya, Le, Gerber,
  Sahai, Williams, and Mahmood]{Chen2024}
Richard~J. Chen, Tong Ding, Ming~Y. Lu, Drew F.~K. Williamson, Guillaume Jaume,
  Andrew~H. Song, Bowen Chen, Andrew Zhang, Daniel Shao, Muhammad Shaban, Mane
  Williams, Lukas Oldenburg, Luca~L. Weishaupt, Judy~J. Wang, Anurag Vaidya,
  Long~Phi Le, Georg Gerber, Sharifa Sahai, Walt Williams, and Faisal Mahmood.
\newblock Towards a general-purpose foundation model for computational
  pathology.
\newblock \emph{Nature Medicine}, 30\penalty0 (3):\penalty0 850--862, March
  2024.
\newblock ISSN 1546-170X.
\newblock \doi{10.1038/s41591-024-02857-3}.

\bibitem[Darcet et~al.(2023)Darcet, Oquab, Mairal, and Bojanowski]{Darcet2023}
Timoth{\'e}e Darcet, Maxime Oquab, Julien Mairal, and Piotr Bojanowski.
\newblock Vision {{Transformers Need Registers}}.
\newblock In \emph{The {{Twelfth International Conference}} on {{Learning
  Representations}}}, October 2023.

\bibitem[Dosovitskiy et~al.(2020)Dosovitskiy, Beyer, Kolesnikov, Weissenborn,
  Zhai, Unterthiner, Dehghani, Minderer, Heigold, Gelly,
  et~al.]{Dosovitskiy2020}
Alexey Dosovitskiy, Lucas Beyer, Alexander Kolesnikov, Dirk Weissenborn,
  Xiaohua Zhai, Thomas Unterthiner, Mostafa Dehghani, Matthias Minderer, Georg
  Heigold, Sylvain Gelly, et~al.
\newblock An image is worth 16x16 words: {{Transformers}} for image recognition
  at scale.
\newblock In \emph{International {{Conference}} on {{Learning Representations}}
  ({{ICLR}} 2020)}, 2020.

\bibitem[Henighan et~al.(2020)Henighan, Kaplan, Katz, Chen, Hesse, Jackson,
  Jun, Brown, Dhariwal, Gray, Hallacy, Mann, Radford, Ramesh, Ryder, Ziegler,
  Schulman, Amodei, and McCandlish]{Henighan2020}
Tom Henighan, Jared Kaplan, Mor Katz, Mark Chen, Christopher Hesse, Jacob
  Jackson, Heewoo Jun, Tom~B. Brown, Prafulla Dhariwal, Scott Gray, Chris
  Hallacy, Benjamin Mann, Alec Radford, Aditya Ramesh, Nick Ryder, Daniel~M.
  Ziegler, John Schulman, Dario Amodei, and Sam McCandlish.
\newblock Scaling {{Laws}} for {{Autoregressive Generative Modeling}}, November
  2020.

\bibitem[Hoffmann et~al.(2022)Hoffmann, Borgeaud, Mensch, Buchatskaya, Cai,
  Rutherford, Casas, Hendricks, Welbl, Clark, Hennigan, Noland, Millican,
  van~den Driessche, Damoc, Guy, Osindero, Simonyan, Elsen, Rae, Vinyals, and
  Sifre]{Hoffmann2022}
Jordan Hoffmann, Sebastian Borgeaud, Arthur Mensch, Elena Buchatskaya, Trevor
  Cai, Eliza Rutherford, Diego de~Las Casas, Lisa~Anne Hendricks, Johannes
  Welbl, Aidan Clark, Tom Hennigan, Eric Noland, Katie Millican, George van~den
  Driessche, Bogdan Damoc, Aurelia Guy, Simon Osindero, Karen Simonyan, Erich
  Elsen, Jack~W. Rae, Oriol Vinyals, and Laurent Sifre.
\newblock Training {{Compute-Optimal Large Language Models}}, March 2022.

\bibitem[Holmes et~al.(1998)Holmes, Hoge, Collins, Woods, Toga, and
  Evans]{Holmes1998}
Colin~J. Holmes, Rick Hoge, Louis Collins, Roger Woods, Arthur~W. Toga, and
  Alan~C. Evans.
\newblock Enhancement of {{MR Images Using Registration}} for {{Signal
  Averaging}}.
\newblock \emph{Journal of Computer Assisted Tomography}, 22\penalty0
  (2):\penalty0 324--333, March 1998.
\newblock ISSN 0363-8715.

\bibitem[Hurlbert(1984)]{Hurlbert1984}
Stuart~H. Hurlbert.
\newblock Pseudoreplication and the {{Design}} of {{Ecological Field
  Experiments}}.
\newblock \emph{Ecological Monographs}, 54\penalty0 (2):\penalty0 187--211,
  1984.
\newblock ISSN 1557-7015.
\newblock \doi{10.2307/1942661}.

\bibitem[Jean et~al.(2018)Jean, Wang, Samar, Azzari, Lobell, and
  Ermon]{Jean2018}
Neal Jean, Sherrie Wang, Anshul Samar, George Azzari, David Lobell, and Stefano
  Ermon.
\newblock {{Tile2Vec}}: {{Unsupervised}} representation learning for spatially
  distributed data, May 2018.

\bibitem[Kaplan et~al.(2020)Kaplan, McCandlish, Henighan, Brown, Chess, Child,
  Gray, Radford, Wu, and Amodei]{Kaplan2020}
Jared Kaplan, Sam McCandlish, Tom Henighan, Tom~B. Brown, Benjamin Chess, Rewon
  Child, Scott Gray, Alec Radford, Jeffrey Wu, and Dario Amodei.
\newblock Scaling {{Laws}} for {{Neural Language Models}}, January 2020.

\bibitem[Lee et~al.(2022)Lee, Ippolito, Nystrom, Zhang, Eck, {Callison-Burch},
  and Carlini]{Lee2022}
Katherine Lee, Daphne Ippolito, Andrew Nystrom, Chiyuan Zhang, Douglas Eck,
  Chris {Callison-Burch}, and Nicholas Carlini.
\newblock Deduplicating {{Training Data Makes Language Models Better}}.
\newblock In Smaranda Muresan, Preslav Nakov, and Aline Villavicencio (eds.),
  \emph{Proceedings of the 60th {{Annual Meeting}} of the {{Association}} for
  {{Computational Linguistics}} ({{Volume}} 1: {{Long Papers}})}, pp.\
  8424--8445, Dublin, Ireland, May 2022. Association for Computational
  Linguistics.
\newblock \doi{10.18653/v1/2022.acl-long.577}.

\bibitem[Lin et~al.(2023)Lin, Yu, Xu, Hu, Xu, and Chen]{Lin2023}
Tiancheng Lin, Zhimiao Yu, Zengchao Xu, Hongyu Hu, Yi~Xu, and Chang-Wen Chen.
\newblock {{SGCL}}: {{Spatial}} guided contrastive learning on whole-slide
  pathological images.
\newblock \emph{Medical Image Analysis}, 89:\penalty0 102845, October 2023.
\newblock ISSN 1361-8415.
\newblock \doi{10.1016/j.media.2023.102845}.

\bibitem[Loshchilov \& Hutter(2018)Loshchilov and Hutter]{Loshchilov2018}
Ilya Loshchilov and Frank Hutter.
\newblock Decoupled {{Weight Decay Regularization}}.
\newblock In \emph{International {{Conference}} on {{Learning
  Representations}}}, September 2018.

\bibitem[Lu et~al.(2024)Lu, Chen, Williamson, Chen, Liang, Ding, Jaume,
  Odintsov, Le, Gerber, Parwani, Zhang, and Mahmood]{Lu2024}
Ming~Y. Lu, Bowen Chen, Drew F.~K. Williamson, Richard~J. Chen, Ivy Liang, Tong
  Ding, Guillaume Jaume, Igor Odintsov, Long~Phi Le, Georg Gerber, Anil~V.
  Parwani, Andrew Zhang, and Faisal Mahmood.
\newblock A visual-language foundation model for computational pathology.
\newblock \emph{Nature Medicine}, 30\penalty0 (3):\penalty0 863--874, March
  2024.
\newblock ISSN 1546-170X.
\newblock \doi{10.1038/s41591-024-02856-4}.

\bibitem[Merker(1983)]{Merker1983}
Bj{\"o}rn Merker.
\newblock Silver staining of cell bodies by means of physical development.
\newblock \emph{Journal of Neuroscience Methods}, 9\penalty0 (3):\penalty0
  235--241, 1983.
\newblock ISSN 0165-0270.
\newblock \doi{10.1016/0165-0270(83)90086-9}.

\bibitem[Muennighoff et~al.(2023)Muennighoff, Rush, Barak, Le~Scao, Tazi,
  Piktus, Pyysalo, Wolf, and Raffel]{Muennighoff2023}
Niklas Muennighoff, Alexander Rush, Boaz Barak, Teven Le~Scao, Nouamane Tazi,
  Aleksandra Piktus, Sampo Pyysalo, Thomas Wolf, and Colin Raffel.
\newblock Scaling {{Data-Constrained Language Models}}.
\newblock In \emph{Advances in {{Neural Information Processing Systems}}},
  volume~36, pp.\  50358--50376. Curran Associates, Inc., 2023.
\newblock \doi{10.52202/075280-2191}.

\bibitem[Oberstrass et~al.(2024)Oberstrass, Muenzing, Niu,
  {Palomero-Gallagher}, Schiffer, Axer, Amunts, and
  Dickscheid]{Oberstrass2024c}
Alexander Oberstrass, Sascha~E.A. Muenzing, Meiqi Niu, Nicola
  {Palomero-Gallagher}, Christian Schiffer, Markus Axer, Katrin Amunts, and
  Timo Dickscheid.
\newblock Self-supervised representation learning for nerve fiber distribution
  patterns in {{3D-PLI}}.
\newblock \emph{Imaging Neuroscience}, 2:\penalty0 imag--2--00351, November
  2024.
\newblock ISSN 2837-6056.
\newblock \doi{10.1162/imag_a_00351}.

\bibitem[Oquab et~al.(2024)Oquab, Darcet, Moutakanni, Vo, Szafraniec, Khalidov,
  Fernandez, Haziza, Massa, {El-Nouby}, Assran, Ballas, Galuba, Howes, Huang,
  Li, Misra, Rabbat, Sharma, Synnaeve, Xu, Jegou, Mairal, Labatut, Joulin, and
  Bojanowski]{Oquab2024}
Maxime Oquab, Timoth{\'e}e Darcet, Th{\'e}o Moutakanni, Huy Vo, Marc
  Szafraniec, Vasil Khalidov, Pierre Fernandez, Daniel Haziza, Francisco Massa,
  Alaaeldin {El-Nouby}, Mahmoud Assran, Nicolas Ballas, Wojciech Galuba,
  Russell Howes, Po-Yao Huang, Shang-Wen Li, Ishan Misra, Michael Rabbat, Vasu
  Sharma, Gabriel Synnaeve, Hu~Xu, Herv{\'e} Jegou, Julien Mairal, Patrick
  Labatut, Armand Joulin, and Piotr Bojanowski.
\newblock {{DINOv2}}: {{Learning Robust Visual Features}} without
  {{Supervision}}, February 2024.

\bibitem[Paszke et~al.(2019)Paszke, Gross, Massa, Lerer, Bradbury, Chanan,
  Killeen, Lin, Gimelshein, Antiga, Desmaison, K{\"o}pf, Yang, DeVito, Raison,
  Tejani, Chilamkurthy, Steiner, Fang, Bai, and Chintala]{Paszke2019}
Adam Paszke, Sam Gross, Francisco Massa, Adam Lerer, James Bradbury, Gregory
  Chanan, Trevor Killeen, Zeming Lin, Natalia Gimelshein, Luca Antiga, Alban
  Desmaison, Andreas K{\"o}pf, Edward Yang, Zach DeVito, Martin Raison, Alykhan
  Tejani, Sasank Chilamkurthy, Benoit Steiner, Lu~Fang, Junjie Bai, and Soumith
  Chintala.
\newblock {{PyTorch}}: {{An Imperative Style}}, {{High-Performance Deep
  Learning Library}}.
\newblock In \emph{33rd {{Conference}} on {{Neural Information Processing
  Systems}} ({{NeurIPS}} 2019)}, 2019.

\bibitem[Pohlen et~al.(2017)Pohlen, Hermans, Mathias, and Leibe]{Pohlen2017}
Tobias Pohlen, Alexander Hermans, Markus Mathias, and Bastian Leibe.
\newblock Full-resolution residual networks for semantic segmentation in street
  scenes.
\newblock In \emph{{{IEEE}} Conference on Computer Vision and Pattern
  Recognition ({{CVPR}}'17)}, 2017.

\bibitem[Schiffer et~al.(2025)Schiffer, Boztoprak, Kropp, Th{\"o}nni{\ss}en,
  Berr, Spitzer, Amunts, and Dickscheid]{Schiffer2025}
Christian Schiffer, Zeynep Boztoprak, Jan-Oliver Kropp, Julia
  Th{\"o}nni{\ss}en, Katia Berr, Hannah Spitzer, Katrin Amunts, and Timo
  Dickscheid.
\newblock {{CytoNet}}: {{A Foundation Model}} for the {{Human Cerebral Cortex}}
  at {{Cellular Resolution}}, October 2025.

\bibitem[Schweiger et~al.(2026)Schweiger, Garthen, Bahn, Chalcraft,
  Schtickzelle, Larsen, and Kreyling]{Schweiger2026}
Andreas~H. Schweiger, Aron Garthen, Michael Bahn, David Chalcraft, Nicolas
  Schtickzelle, Klaus~Steenberg Larsen, and J{\"u}rgen Kreyling.
\newblock How to optimally allocate sampling effort in experimental ecology.
\newblock \emph{Scientific Reports}, 16\penalty0 (1):\penalty0 6503, February
  2026.
\newblock ISSN 2045-2322.
\newblock \doi{10.1038/s41598-026-38541-4}.

\bibitem[Sorscher et~al.(2023)Sorscher, Geirhos, Shekhar, Ganguli, and
  Morcos]{Sorscher2023}
Ben Sorscher, Robert Geirhos, Shashank Shekhar, Surya Ganguli, and Ari~S.
  Morcos.
\newblock Beyond neural scaling laws: Beating power law scaling via data
  pruning, April 2023.

\bibitem[Tirumala et~al.(2023)Tirumala, Simig, Aghajanyan, and
  Morcos]{Tirumala2023}
Kushal Tirumala, Daniel Simig, Armen Aghajanyan, and Ari Morcos.
\newblock D4: {{Improving LLM}} pretraining via document de-duplication and
  diversification.
\newblock In A.~Oh, T.~Naumann, A.~Globerson, K.~Saenko, M.~Hardt, and
  S.~Levine (eds.), \emph{Advances in Neural Information Processing Systems},
  volume~36, pp.\  53983--53995. Curran Associates, Inc., 2023.
\newblock \doi{10.52202/075280-2348}.

\bibitem[Vorontsov et~al.(2024)Vorontsov, Bozkurt, Casson, Shaikovski,
  Zelechowski, Severson, Zimmermann, Hall, Tenenholtz, Fusi, Yang, Mathieu,
  {van Eck}, Lee, Viret, Robert, Wang, Kunz, Lee, Bernhard, Godrich, Oakley,
  Millar, Hanna, Wen, Retamero, Moye, Yousfi, Kanan, Klimstra, Rothrock, Liu,
  and Fuchs]{Vorontsov2024}
Eugene Vorontsov, Alican Bozkurt, Adam Casson, George Shaikovski, Michal
  Zelechowski, Kristen Severson, Eric Zimmermann, James Hall, Neil Tenenholtz,
  Nicolo Fusi, Ellen Yang, Philippe Mathieu, Alexander {van Eck}, Donghun Lee,
  Julian Viret, Eric Robert, Yi~Kan Wang, Jeremy~D. Kunz, Matthew C.~H. Lee,
  Jan~H. Bernhard, Ran~A. Godrich, Gerard Oakley, Ewan Millar, Matthew Hanna,
  Hannah Wen, Juan~A. Retamero, William~A. Moye, Razik Yousfi, Christopher
  Kanan, David~S. Klimstra, Brandon Rothrock, Siqi Liu, and Thomas~J. Fuchs.
\newblock A foundation model for clinical-grade computational pathology and
  rare cancers detection.
\newblock \emph{Nature Medicine}, 30\penalty0 (10):\penalty0 2924--2935,
  October 2024.
\newblock ISSN 1546-170X.
\newblock \doi{10.1038/s41591-024-03141-0}.

\bibitem[Wang et~al.(2026)Wang, Ren, Zhang, Neskovic, Bhattad, Xie, and
  Yuille]{Wang2026}
Feng Wang, Sucheng Ren, Tiezheng Zhang, Predrag Neskovic, Anand Bhattad, Cihang
  Xie, and Alan Yuille.
\newblock {{ViT-5}}: {{Vision Transformers}} for {{The Mid-2020s}}, February
  2026.

\bibitem[Wu et~al.(2025)Wu, Zhang, Ru, Dang, Lao, Yu, Luo, Zhu, Sun, Zhang,
  Zhu, Wang, Yang, Chen, Zhang, and Li]{Wu2025}
Kang Wu, Yingying Zhang, Lixiang Ru, Bo~Dang, Jiangwei Lao, Lei Yu, Junwei Luo,
  Zifan Zhu, Yue Sun, Jiahao Zhang, Qi~Zhu, Jian Wang, Ming Yang, Jingdong
  Chen, Yongjun Zhang, and Yansheng Li.
\newblock A semantic-enhanced multi-modal remote sensing foundation model for
  {{Earth}} observation.
\newblock \emph{Nature Machine Intelligence}, 7\penalty0 (8):\penalty0
  1235--1249, August 2025.
\newblock ISSN 2522-5839.
\newblock \doi{10.1038/s42256-025-01078-8}.

\bibitem[Xu et~al.(2024)Xu, Usuyama, Bagga, Zhang, Rao, Naumann, Wong, Gero,
  Gonz{\'a}lez, Gu, Xu, Wei, Wang, Ma, Wei, Yang, Li, Gao, Rosemon, Bower, Lee,
  Weerasinghe, Wright, Robicsek, Piening, Bifulco, Wang, and Poon]{Xu2024}
Hanwen Xu, Naoto Usuyama, Jaspreet Bagga, Sheng Zhang, Rajesh Rao, Tristan
  Naumann, Cliff Wong, Zelalem Gero, Javier Gonz{\'a}lez, Yu~Gu, Yanbo Xu,
  Mu~Wei, Wenhui Wang, Shuming Ma, Furu Wei, Jianwei Yang, Chunyuan Li,
  Jianfeng Gao, Jaylen Rosemon, Tucker Bower, Soohee Lee, Roshanthi
  Weerasinghe, Bill~J. Wright, Ari Robicsek, Brian Piening, Carlo Bifulco,
  Sheng Wang, and Hoifung Poon.
\newblock A whole-slide foundation model for digital pathology from real-world
  data.
\newblock \emph{Nature}, 630\penalty0 (8015):\penalty0 181--188, June 2024.
\newblock ISSN 1476-4687.
\newblock \doi{10.1038/s41586-024-07441-w}.

\bibitem[Xu et~al.(2023)Xu, Wang, Fang, Luo, Chen, Wan, Wang, Tang, Xue, Li,
  Lin, and Qu]{Xu2023}
Hao Xu, Shuyan Wang, Minghao Fang, Songwen Luo, Chunpeng Chen, Siyuan Wan,
  Rirui Wang, Meifang Tang, Tian Xue, Bin Li, Jun Lin, and Kun Qu.
\newblock {{SPACEL}}: Deep learning-based characterization of spatial
  transcriptome architectures.
\newblock \emph{Nature Communications}, 14\penalty0 (1):\penalty0 7603,
  November 2023.
\newblock ISSN 2041-1723.
\newblock \doi{10.1038/s41467-023-43220-3}.

\bibitem[Yagis et~al.(2021)Yagis, Atnafu, {Garc{\'i}a Seco de Herrera}, Marzi,
  Scheda, Giannelli, Tessa, Citi, and Diciotti]{Yagis2021}
Ekin Yagis, Selamawet~Workalemahu Atnafu, Alba {Garc{\'i}a Seco de Herrera},
  Chiara Marzi, Riccardo Scheda, Marco Giannelli, Carlo Tessa, Luca Citi, and
  Stefano Diciotti.
\newblock Effect of data leakage in brain {{MRI}} classification using {{2D}}
  convolutional neural networks.
\newblock \emph{Scientific Reports}, 11\penalty0 (1):\penalty0 22544, November
  2021.
\newblock ISSN 2045-2322.
\newblock \doi{10.1038/s41598-021-01681-w}.

\bibitem[Zimmermann et~al.(2024)Zimmermann, Vorontsov, Viret, Casson,
  Zelechowski, Shaikovski, Tenenholtz, Hall, Klimstra, Yousfi, Fuchs, Fusi,
  Liu, and Severson]{Zimmermann2024}
Eric Zimmermann, Eugene Vorontsov, Julian Viret, Adam Casson, Michal
  Zelechowski, George Shaikovski, Neil Tenenholtz, James Hall, David Klimstra,
  Razik Yousfi, Thomas Fuchs, Nicolo Fusi, Siqi Liu, and Kristen Severson.
\newblock Virchow2: {{Scaling Self-Supervised Mixed Magnification Models}} in
  {{Pathology}}, November 2024.

\end{thebibliography}
\bibliographystyle{preprint}

\newpage
\appendix
\crefalias{section}{appendix}
\crefalias{subsection}{appendix}

\section{Data and task}\label{app:data}

\subsection{Subjects, tissue processing, and consent}\label{app:tissue}

The dataset comprises 21 postmortem human brains.
Subject characteristics and the tissue-processing and digitization procedures are described in \citet{Amunts2020} and \citet{Schiffer2025}.
Ethics approval and the body-donor consent procedure are stated in the ethics statement of the main text.
Brains were chemically fixed, paraffin-embedded, sectioned at $20\,\mu\mathrm{m}$, stained for neuronal cell bodies with a modified silver protocol~\citep{Merker1983}, and digitized at $1\,\mu\mathrm{m}$ per pixel.
Approximately every fifteenth physical section was retained, yielding 300--506 sections per brain and \num{11644201} sampling locations in total.
Sixteen brains were cut coronally and five horizontally.
Cutting plane was recorded for each subject and used to stratify subject subsets.
\textbf{E4} is restricted to the 13 coronal subjects in the pretraining pool because its block-based coverage manipulation requires a common sectioning axis across subjects.
\Cref{tab:s_subjects} gives the per-subject inventory.

\begin{table}[t]
	\caption{\textbf{Dataset overview.}
		For each subject, we report the number of sampling locations, retained sections, annotated sections, cutting plane, and role in the study.
		Subjects are identified by numeric IDs.
	}
	\label{tab:s_subjects}
	\begin{center}
		\begin{tabular}{@{}lrrrll@{}}
	\toprule
	\textbf{subject} & \textbf{locations} & \textbf{sections} & \textbf{annotated} & \textbf{plane} & \textbf{role} \\
	\midrule
	B01              & \num{644224}       & 468               & 332                & coronal        & pretraining   \\
	B02              & \num{522551}       & 462               & 253                & coronal        & pretraining   \\
	B03              & \num{610975}       & 506               & 298                & coronal        & pretraining   \\
	B04              & \num{570373}       & 475               & 372                & coronal        & transfer      \\
	B05              & \num{461204}       & 451               & 355                & coronal        & pretraining   \\
	B06              & \num{571496}       & 469               & 385                & coronal        & pretraining   \\
	B07              & \num{597666}       & 487               & 378                & coronal        & held out      \\
	B08              & \num{514651}       & 451               & 343                & coronal        & pretraining   \\
	B09              & \num{655532}       & 476               & 353                & coronal        & held out      \\
	B10              & \num{533815}       & 478               & 346                & coronal        & pretraining   \\
	B11              & \num{557567}       & 499               & 329                & coronal        & transfer      \\
	B12              & \num{496407}       & 465               & 221                & coronal        & pretraining   \\
	B13              & \num{499678}       & 431               & 264                & coronal        & pretraining   \\
	B14              & \num{500528}       & 460               & 259                & coronal        & pretraining   \\
	B15              & \num{607156}       & 387               & 105                & horizontal     & pretraining   \\
	B16              & \num{566503}       & 300               & 83                 & horizontal     & transfer      \\
	B17              & \num{619245}       & 322               & 158                & horizontal     & pretraining   \\
	B18              & \num{532668}       & 339               & 138                & horizontal     & pretraining   \\
	B20              & \num{556888}       & 453               & 451                & coronal        & held out      \\
	B21              & \num{668055}       & 492               & 216                & coronal        & pretraining   \\
	B23              & \num{357019}       & 322               & 0                  & horizontal     & pretrain only \\
	\bottomrule
\end{tabular}

	\end{center}
\end{table}

\subsection{Sampling-location generation}\label{app:sampling}

Candidate locations were generated independently within each section from gray- and white-matter masks using repeated Euclidean distance transforms.
At each iteration, the procedure selected the point with the largest distance to the nearest previously selected point or tissue boundary, excluded a $2\times2\,\mathrm{mm}$ square around it, and repeated until the maximum remaining distance fell below $300\,\mu\mathrm{m}$ in gray matter or $500\,\mu\mathrm{m}$ in white matter.
Gray- and white-matter masks were generated as described in \citet{Schiffer2025} using thresholding and intensity transformations.
Because sampling was performed independently within each section, the procedure did not require surface reconstruction or section-to-section alignment and could sample cortical and subcortical gray matter as well as white matter.
The procedure yielded \num{11644201} locations, approximately \num{554000} per subject.
Each location was mapped to the Colin27 reference space~\citep{Holmes1998} following \citet{Schiffer2025}.
The resulting common coordinate system defines Euclidean distances between samples from different subjects.

\subsection{Downstream task and evaluation semantics}\label{app:probe}

The probe predicts one of 191 cytoarchitectonic areas from a class vocabulary fixed before evaluation and derived from the Julich Brain atlas~\citep{Amunts2020}.
Probe-training samples are capped at \num{1000} per class without replacement; capped classes are filled round-robin across subjects, with a fixed-seed random order within each subject.
The probe is a single affine layer mapping the frozen encoder representation to the 191 classes and is trained with categorical cross-entropy.
Encoder features are passed to the classifier without normalization or other transformation.
The probe is optimized with AdamW at a constant learning rate of $10^{-3}$ and weight decay $0.05$, using a global batch size of \num{2048} across 64 GH200 accelerators for \num{4710} steps.
The fixed step budget keeps probe optimization effort independent of training-pool size.
No physical section contributes to both probe training and evaluation.
Macro-F1 is computed separately for each evaluation subject over the classes present in that subject---137, 171, and 172 of the 191 classes for the three primary unseen subjects, respectively---and then averaged across subjects.
Three subjects are excluded from all controlled pretraining runs and from probe training and form the primary unseen evaluation set.
Three additional subjects are excluded from probe training but remain eligible for pretraining and define the transfer subjects.
The remaining annotated subjects contribute to probe training, with separate sections held out for evaluation.
Generalization level is therefore defined for each model--subject pair.

\section{Models, training, and compute}\label{app:models}

\subsection{Architecture and optimization}\label{app:arch}

Encoders use the modernized ViT-5 architecture of \citet{Wang2026}, with SwiGLU feed-forward blocks of hidden size $8/3$ times the embedding dimension, bias-free QKV projections, no stochastic depth, and four register tokens~\citep{Darcet2023}.
Each $2048\times2048$ input is divided into a $64\times64$ grid of $32\times32$-pixel patches, yielding \num{4096} patch tokens and a sequence length of \num{4101} after adding the class token and four register tokens; the class token provides the encoder representation.

\begin{table}[t]
	\caption{
		\textbf{Encoder dimensions.}
		Depth, width, and head count follow the standard ViT variants, while token grid and register count are identical across capacities.
	}
	\label{tab:s_arch}
	\begin{center}
		\begin{tabular}{@{}lrrrr@{}}
			\toprule
			\textbf{Variant} & \textbf{Width} & \textbf{Heads} & \textbf{Layers} & \textbf{Accelerators} \\
			\midrule
			ViT-B            & 768            & 12             & 12              & 64                    \\
			ViT-L            & 1024           & 16             & 24              & 128                   \\
			ViT-H            & 1280           & 16             & 32              & 256                   \\
			\bottomrule
		\end{tabular}
	\end{center}
\end{table}

All models are trained with AdamW~\citep{Loshchilov2018} at a constant learning rate of $10^{-4}$, without warmup or decay, using weight decay $0.05$ for weights and $0$ for normalization parameters and biases.
Pretraining and linear-probe training use automatic mixed precision with bfloat16 autocast.
The global batch size is fixed at \num{2048} for all capacities.
Larger encoders use more accelerators while retaining the same global batch size and step budget, separating model capacity from training duration.
Representations are gathered across all accelerators before computing SpatialNCE, so the objective is evaluated over the full global batch.
The projection head maps the class-token representation to $z_i$ using a two-layer MLP consisting of a linear layer to the encoder width, RMS normalization, GELU, and a bias-free linear layer to 256 dimensions.
The resulting vector is $L_2$-normalized before entering SpatialNCE.
The projection head is discarded after pretraining, and downstream evaluation uses the encoder class-token representation.

Augmentation follows the CytoNet protocol of \citet{Schiffer2025}.
Patches are rotated by $\theta\sim U[-\pi,+\pi]$, mirrored with probability $0.5$, and translated in a random direction by up to $0.2\,\mathrm{mm}$.
Pixel intensities $x\in[0,1]$ are transformed using unbiased gamma augmentation~\citep{Pohlen2017} as
$\alpha x^\gamma+\beta$,
with $\alpha\sim U[0.9,1.1]$, $\beta\sim U[-0.1,+0.1]$,
\[
	\gamma=\frac{\log\left(0.5+2^{-0.5}Z\right)}{\log\left(0.5-2^{-0.5}Z\right)},
\]
and $Z\sim U[-0.05,+0.05]$.
With probability $0.25$ each, a patch is blurred by an isotropic Gaussian with $\sigma\sim U[0.125,1.0]$ or sharpened as $x+\delta(x-G_{\sigma_u}(x))$, with $\sigma_u\sim U[0.125,2.0]$ and $\delta\sim U[0.5,1.5]$.
Patches are standardized to $[-1,1]$ after augmentation.

\subsection{Compute effort}\label{app:compute}

Pretraining consumed \num{315973} accelerator-hours over \num{7320000} optimization steps and $\num{94200} \times10^{18}$ FLOPs (\num{94200} exaFLOPs).
The 124 linear probes consumed an additional \num{32751} accelerator-hours, giving \num{348724} accelerator-hours in total.
FLOPs are computed analytically from dense matrix operations at \num{4101} tokens, counting two operations per multiply--add and excluding normalization, softmax, activation functions, and optimizer updates.
Superseded training attempts, data loading, and prediction jobs are excluded from these totals.

\subsection{Software and execution environment}\label{app:env}

All runs were executed on the JUPITER supercomputer at the Jülich Supercomputing Centre.
Its accelerated partition uses NVIDIA GH200 Grace--Hopper superchips with four accelerators per node.
Pretraining used 16 nodes for ViT-B, 32 nodes for ViT-L, and 64 nodes for ViT-H, corresponding to 64, 128, and 256 accelerators, respectively.
The global pretraining batch size was fixed at \num{2048} across all capacities.
Linear probes used 16 nodes with 64 accelerators.
All pretraining and probe jobs ran under SLURM in the same Apptainer image, providing a common software environment across runs.
Training used PyTorch~\citep{Paszke2019} and distributed data parallel (DDP) training.
Statistical analysis and figure generation were performed in a separate environment using NumPy, pandas, SciPy, and statsmodels.

\section{Statistical analysis}\label{app:stats}

\subsection{Derivation of the two detection thresholds}\label{app:stats_sensitivity}

We derive separate minimum detectable effects (MDEs) for contrasts between configurations trained on different pretraining-subject subsets and contrasts between configurations trained on the same subset.
For configuration $c$ evaluated on subject $d$, we write the fitted mixed-effects model as
\[
	Y_{cd}
	=
	\mu_c + u_{s(c)} + v_d + e_{cd},
\]
where $\mu_c$ is the fixed-effect contribution for configuration $c$, $u_{s(c)}$ is the random intercept for its pretraining-subject subset $s(c)$, $v_d$ is the evaluation-subject random intercept, and $e_{cd}$ is the residual.
The fitted variance components relevant to these contrasts are $\hat{\sigma}_u^2=0.00150$ and $\hat{\sigma}_e^2=0.00525$.

Every configuration is evaluated on the same held-out subjects, so configuration contrasts are paired by evaluation subject.
For two configurations $c_1$ and $c_2$ evaluated on the same subject $d$,
\[
	Y_{c_1d}-Y_{c_2d}
	=
	(\mu_{c_1}-\mu_{c_2})
	+
	(u_{s(c_1)}-u_{s(c_2)})
	+
	(e_{c_1d}-e_{c_2d}).
\]
The evaluation-subject effect $v_d$ cancels in this paired contrast.

For an across-subset contrast, the fitted model treats $u_{s(c_1)}$ and $u_{s(c_2)}$ as independent, giving
\[
	\operatorname{Var}\!\left(u_{s(c_1)}-u_{s(c_2)}\right)
	=
	2\sigma_u^2.
\]
This subset-level contribution is shared across evaluation subjects and is unchanged when their contrasts are averaged.
The residuals are independent across evaluation subjects under the fitted model, so averaging over $n_d$ subjects reduces the residual contribution to $2\sigma_e^2/n_d$.
Writing $\bar{\Delta}$ for the configuration contrast averaged over the $n_d$ evaluation subjects gives
\[
	\operatorname{Var}(\bar{\Delta}_{\mathrm{across}})
	=
	2\sigma_u^2+\frac{2\sigma_e^2}{n_d}.
\]

For a within-subset contrast, the pretraining-subject subset effect also cancels.
The variance of the averaged contrast is then
\[
	\operatorname{Var}(\bar{\Delta}_{\mathrm{within}})
	=
	\frac{2\sigma_e^2}{n_d}.
\]

With $n_d=3$ evaluation subjects, the corresponding standard errors are
\[
	\operatorname{SE}_{\mathrm{across}}
	=
	\sqrt{2\hat{\sigma}_u^2+\frac{2\hat{\sigma}_e^2}{3}},
	\qquad
	\operatorname{SE}_{\mathrm{within}}
	=
	\sqrt{\frac{2\hat{\sigma}_e^2}{3}}.
\]

Using a normal approximation for a two-sided test with $\alpha=0.05$ and $80\%$ power, we define
\[
	\mathrm{MDE}
	=
	\left(z_{1-\alpha/2}+z_{0.80}\right)\operatorname{SE}
	=
	\left(z_{0.975}+z_{0.80}\right)\operatorname{SE},
\]
where $z_{0.975}+z_{0.80}\approx2.80$.
This gives an across-subset MDE of $0.226$ logit units and a within-subset MDE of $0.166$ logit units.
Because the logit transformation is nonlinear, the corresponding macro-F1 difference depends on the baseline score $S$.
At $S=0.15$, the across- and within-subset MDEs correspond to approximately $0.029$ and $0.021$ macro-F1, respectively.
At $S=0.21$, they correspond to approximately $0.038$ and $0.028$ macro-F1, respectively.
At $S=0.30$, they correspond to approximately $0.047$ and $0.035$ macro-F1, respectively.
These MDEs apply to pairwise contrasts between two configurations.

For the aggregated familiarity analysis in \cref{sec:r1}, the across-subset MDE is used only as a reference effect-size scale.
All 17 subject-level familiarity contrasts exceed this value.

\subsection{Collinearity among the fixed effects}\label{app:vif}

The full-grid model fits a common additive specification across experiments with different manipulations.
Several fixed effects are correlated by design, particularly $N_{\mathrm{unique}}$, the coverage indicators $\rho$, and the positive weight $M$.
We quantify linear collinearity for each continuous predictor using variance inflation factors (VIFs) computed from the encoded fixed-effect design matrix.

For continuous predictor $j$,
\[
	\mathrm{VIF}_j=\frac{1}{1-R_j^2},
\]
where $R_j^2$ is obtained by regressing predictor $j$ on all remaining columns of the fixed-effect design matrix.

\begin{table}[t]
	\caption{\textbf{Variance inflation factors for the continuous predictors in the full-grid model.}
		All VIFs are below $2$.
	}
	\label{tab:s_vif}
	\begin{center}
		\begin{tabular}{@{}llr@{}}
	\toprule
	\textbf{fit} & \textbf{term} & \textbf{VIF} \\
	\midrule
	full grid    & $\beta_N$     & 1.92         \\
	full grid    & $\beta_B$     & 1.73         \\
	full grid    & $\beta_C$     & 1.00         \\
	full grid    & $\beta_M$     & 1.14         \\
	fixed sample & $\beta_B$     & 1.37         \\
	fixed sample & $\beta_M$     & 1.37         \\
	\bottomrule
\end{tabular}

	\end{center}
\end{table}

The largest VIF is $1.92$ for $\log_2 N_{\mathrm{unique}}$, indicating limited linear dependence among the encoded fixed effects (\cref{tab:s_vif}).
In \textbf{E5}, compute is varied through optimization steps while data composition is held fixed, providing direct variation in $C$ independent of the data-allocation factors.
The VIFs quantify linear dependence within the encoded design matrix but do not test whether a common additive model is an adequate specification across the heterogeneous experiments.

\subsection{Sensitivity of the capacity interaction}\label{app:power}

\begin{figure}[t]
	\begin{center}
		\includegraphics[width=0.5\textwidth]{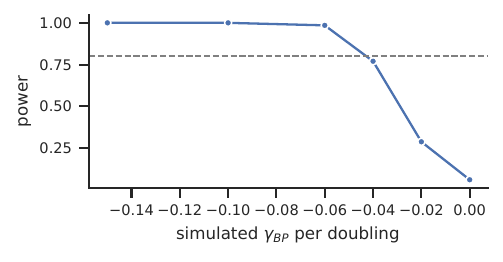}
	\end{center}
	\caption{\textbf{Simulated power for detecting a capacity-by-subject-count interaction $\gamma_{BP}$ under \textbf{E1}.}
		Power is shown as a function of the imposed interaction magnitude and residual variance.
		At the fitted residual variance, approximately $80\%$ power is reached for an interaction of $0.05$ logit units per doubling of subject count.
	}
	\label{fig:s_power}
\end{figure}

To quantify sensitivity to a capacity-by-subject-count interaction, we simulated outcomes under the realized design of \textbf{E1}~(\cref{fig:s_power}).
The design matrix was fixed, while the imposed interaction coefficient $\gamma_{BP}$ and residual variance were varied.
All remaining model parameters were fixed to their estimates from the observed data.
For each parameter setting, we generated repeated synthetic datasets and refit the fixed-sample mixed-effects model.
Power was estimated as the fraction of fits whose 95\% confidence interval for $\gamma_{BP}$ excluded zero.

At the fitted residual variance, interactions smaller than approximately $0.05$ logit units per doubling of subject count are not reliably detectable under \textbf{E1}.
The absence of a detected capacity-by-subject-count interaction therefore excludes only effects above this approximate sensitivity threshold.

\subsection{Pre-specification and multiplicity}\label{app:prespec}

The capacity-by-subject-count interaction in \textbf{E1} was designated as a primary analysis before inspection of outcomes.
The contrast between the 16-block and every-10th-section conditions in \textbf{E4}, which approximately matches both $N_{\mathrm{unique}}$ and $M$, was also designated as primary.
Individual localized-coverage contrasts in \textbf{E4}, subject-gating contrasts in \textbf{E7}, familiarity comparisons, and reference-encoder comparisons were treated as exploratory.
Exploratory contrasts are reported with unadjusted 95\% confidence intervals, without multiplicity correction.
No primary conclusion relies solely on whether an exploratory confidence interval excludes zero.

\section{Additional results}\label{app:results}

\subsection{Coverage geometry and subject gating}\label{app:coverage}

\begin{table}[t]
	\caption{\textbf{Geometry of the coverage conditions $\rho$ in \textbf{E4} at $B=13$, averaged across the 13 coronal subjects.}
		All conditions use the same sample budget but differ in anatomical extent and spatial fragmentation.
	}
	\label{tab:s_geometry}
	\begin{center}
		\begin{tabular}{@{}lrrr@{}}
	\toprule
	\textbf{condition $\rho$} & \textbf{sections} & \textbf{A--P span (mm)} & \textbf{largest gap (mm)} \\
	\midrule
	all                       & 467               & 180                     & 1.0                       \\
	every 10th                & 47                & 180                     & 5.2                       \\
	16 blocks                 & 78                & 175                     & 11.9                      \\
	4 blocks                  & 52                & 142                     & 42.5                      \\
	loc.\ posterior           & 97                & 37                      & 0.5                       \\
	loc.\ center              & 31                & 11                      & 0.5                       \\
	loc.\ anterior            & 88                & 34                      & 0.8                       \\
	\bottomrule
\end{tabular}

	\end{center}
\end{table}

\Cref{fig:s_coverage} summarizes the performance and positive-weight changes induced by the coverage manipulation in \textbf{E4} and subject gating in \textbf{E7}.
\Cref{tab:s_geometry} details the section count, anatomical extent, and fragmentation of the \textbf{E4} coverage conditions at $B=13$.
The three localized conditions in \textbf{E4} match in sample count but differ in the number and anatomical distribution of sampled sections.
Central sections contain substantially more tissue area, so 31 such sections provide approximately as many sampling locations as 97 posterior or 88 anterior sections.
The localized-center condition therefore has the smallest anterior--posterior extent among the three localized conditions.
Its positive weight $M$ is not the largest, because $M$ depends on the full three-dimensional distribution of pairwise distances.

\subsection{Composition of the SpatialNCE objective}\label{app:kernel}

\begin{figure}[t]
	\begin{center}
		\includegraphics[width=\textwidth]{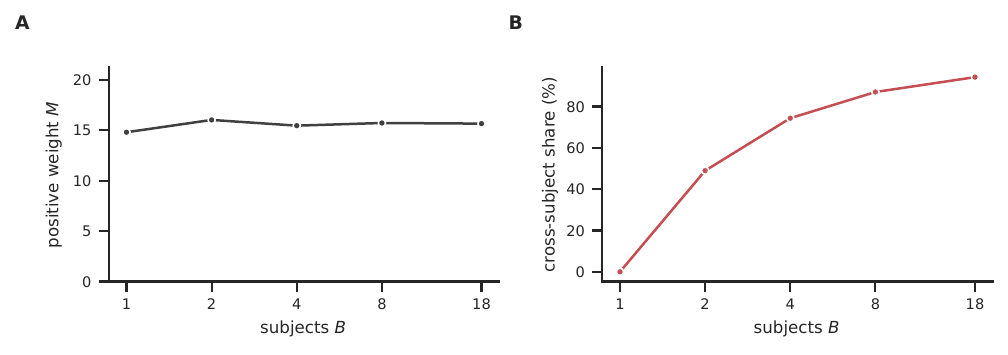}
	\end{center}
	\caption{\textbf{Positive weight across \textbf{E1}.}
		\textbf{A}: Total positive weight $M$ remains approximately constant as subject count increases from one to 18.
		\textbf{B}: The fraction of $M$ contributed by cross-subject pairs increases from zero to $94.2\%$ over the same range.
		At fixed $N_{\mathrm{unique}}$, \textbf{E1} therefore preserves the overall positive weight while changing the composition of contributing pairs.
	}
	\label{fig:s_kernel}
\end{figure}

\Cref{fig:s_kernel} shows the total positive weight $M$ and its cross-subject contribution across \textbf{E1}, while \cref{tab:s_kernel} reports both quantities for each pretraining configuration.
Realized location counts in \textbf{E1} range from \num{448861} to \num{450825}, a spread below $0.5\%$, and the number of sample presentations is identical across conditions.
Across ungated configurations outside \textbf{E4}, $M$ ranges only from $14.5$ to $17.6$ despite a 224-fold range in $N_{\mathrm{unique}}$ and variation across all tested subject counts.
\textbf{E4} increases $M$ as samples become more spatially concentrated, reaching $17.9$ and $28.1$ for the fragmented conditions and $53$--$115$ for the localized conditions.
\textbf{E7} decreases $M$ by removing cross-subject contributions, leaving $M=8.34$ at $B=2$ and $M=0.92$ at $B=18$, approximately 17-fold below the corresponding ungated value at $B=18$.

\begin{table}[t]
	\caption{\textbf{Positive weight $M$ and its cross-subject contribution for every pretraining configuration, averaged across subject subsets at each level.}
		$M$ is the covariate $M_c$ used in the mixed-effects model and is computed from the configuration before training.
	}
	\label{tab:s_kernel}
	\begin{center}
		\begin{tabular}{@{}rlcrrrr@{}}
	\toprule
	\textbf{$B$} & \textbf{sampling} & \textbf{gated} & \textbf{subsets} & \textbf{$N_{\mathrm{unique}}$} & \textbf{$M$} & \textbf{cross-subject (\%)} \\
	\midrule
	1            & every 10th        & \ding{55}      & 5                & \num{51709}                    & 15.23        & 0.0                         \\
	2            & every 10th        & \ding{55}      & 4                & \num{108881}                   & 16.31        & 49.0                        \\
	4            & every 10th        & \ding{55}      & 5                & \num{219266}                   & 15.95        & 74.0                        \\
	8            & every 10th        & \ding{55}      & 2                & \num{438674}                   & 15.85        & 86.9                        \\
	9            & every 10th        & \ding{55}      & 1                & \num{472504}                   & 15.66        & 88.4                        \\
	13           & every 10th        & \ding{55}      & 1                & \num{701132}                   & 16.15        & 92.0                        \\
	18           & every 10th        & \ding{55}      & 1                & \num{962026}                   & 15.73        & 94.1                        \\
	1            & every 3rd         & \ding{55}      & 1                & \num{164378}                   & 14.65        & 0.0                         \\
	9            & every 3rd         & \ding{55}      & 1                & \num{1579631}                  & 15.59        & 88.4                        \\
	18           & every 3rd         & \ding{55}      & 1                & \num{3225383}                  & 15.74        & 94.2                        \\
	1            & 4 blocks          & \ding{55}      & 1                & \num{49036}                    & 28.08        & 0.0                         \\
	4            & 4 blocks          & \ding{55}      & 1                & \num{223273}                   & 27.11        & 70.4                        \\
	13           & 4 blocks          & \ding{55}      & 1                & \num{706946}                   & 24.70        & 90.1                        \\
	13           & 16 blocks         & \ding{55}      & 1                & \num{711367}                   & 17.94        & 91.7                        \\
	1            & all               & \ding{55}      & 5                & \num{532596}                   & 15.23        & 0.0                         \\
	2            & all               & \ding{55}      & 4                & \num{1108908}                  & 16.33        & 49.1                        \\
	4            & all               & \ding{55}      & 5                & \num{2238444}                  & 15.94        & 74.0                        \\
	8            & all               & \ding{55}      & 2                & \num{4481222}                  & 15.77        & 86.9                        \\
	9            & all               & \ding{55}      & 1                & \num{4816705}                  & 15.59        & 88.4                        \\
	13           & all               & \ding{55}      & 1                & \num{7151524}                  & 16.12        & 92.0                        \\
	18           & all               & \ding{55}      & 3                & \num{9942251}                  & 15.68        & 94.1                        \\
	1            & fixed budget      & \ding{55}      & 2                & \num{450358}                   & 14.82        & 0.0                         \\
	2            & fixed budget      & \ding{55}      & 2                & \num{450354}                   & 16.04        & 48.9                        \\
	4            & fixed budget      & \ding{55}      & 2                & \num{449824}                   & 15.47        & 74.4                        \\
	8            & fixed budget      & \ding{55}      & 2                & \num{449761}                   & 15.74        & 87.0                        \\
	18           & fixed budget      & \ding{55}      & 3                & \num{449192}                   & 15.67        & 94.2                        \\
	1            & loc.\ anterior    & \ding{55}      & 1                & \num{48872}                    & 115.43       & 0.0                         \\
	4            & loc.\ anterior    & \ding{55}      & 1                & \num{221468}                   & 107.81       & 72.9                        \\
	13           & loc.\ anterior    & \ding{55}      & 1                & \num{708859}                   & 105.58       & 91.5                        \\
	1            & loc.\ center      & \ding{55}      & 1                & \num{50094}                    & 64.96        & 0.0                         \\
	4            & loc.\ center      & \ding{55}      & 1                & \num{223022}                   & 63.40        & 69.4                        \\
	13           & loc.\ center      & \ding{55}      & 1                & \num{712904}                   & 53.11        & 89.5                        \\
	1            & loc.\ posterior   & \ding{55}      & 1                & \num{49110}                    & 98.57        & 0.0                         \\
	4            & loc.\ posterior   & \ding{55}      & 1                & \num{221797}                   & 106.65       & 73.0                        \\
	13           & loc.\ posterior   & \ding{55}      & 1                & \num{707964}                   & 99.05        & 91.4                        \\
	18           & every 10th        & \ding{51}      & 1                & \num{962026}                   & 0.92         & 0.0                         \\
	2            & all               & \ding{51}      & 1                & \num{1090235}                  & 8.34         & 0.0                         \\
	4            & all               & \ding{51}      & 1                & \num{2233423}                  & 3.97         & 0.0                         \\
	8            & all               & \ding{51}      & 1                & \num{4356763}                  & 2.07         & 0.0                         \\
	18           & all               & \ding{51}      & 1                & \num{9834115}                  & 0.92         & 0.0                         \\
	\bottomrule
\end{tabular}

	\end{center}
\end{table}

\subsection{Spatial localization of the familiarity benefit}\label{app:decay}

\begin{figure}[t]
	\begin{center}
		\includegraphics[width=0.5\textwidth]{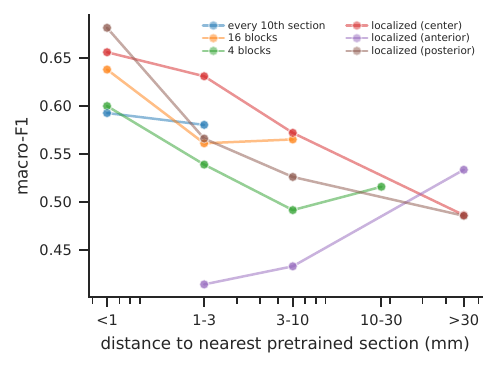}
	\end{center}
	\caption{\textbf{Macro-F1 on held-out sections from familiar subjects as a function of distance to the nearest section seen during pretraining in the $B=13$ runs of \textbf{E4}.}
		Each line corresponds to one coverage condition $\rho$.
		Macro-F1 decreases with distance from the nearest pretraining section in five of the six coverage conditions.
	}
	\label{fig:s_decay}
\end{figure}

\Cref{fig:s_decay} resolves the familiarity effect spatially across the $B=13$ conditions of \textbf{E4}.
Macro-F1 decreases with distance to the nearest section seen during pretraining in five of the six coverage conditions.
This pattern is consistent with a familiarity benefit that is strongest near tissue encountered during pretraining.
The analysis is confounded by spatial variation in the difficulty and class composition of cytoarchitectonic areas.
For example, performance in the anterior-localized condition increases with distance from the pretraining sections, which may reflect higher classification performance for areas located farther posteriorly.

\subsection{Pretraining behavior}\label{app:training}

\begin{figure}[t]
	\begin{center}
		\includegraphics[width=\textwidth]{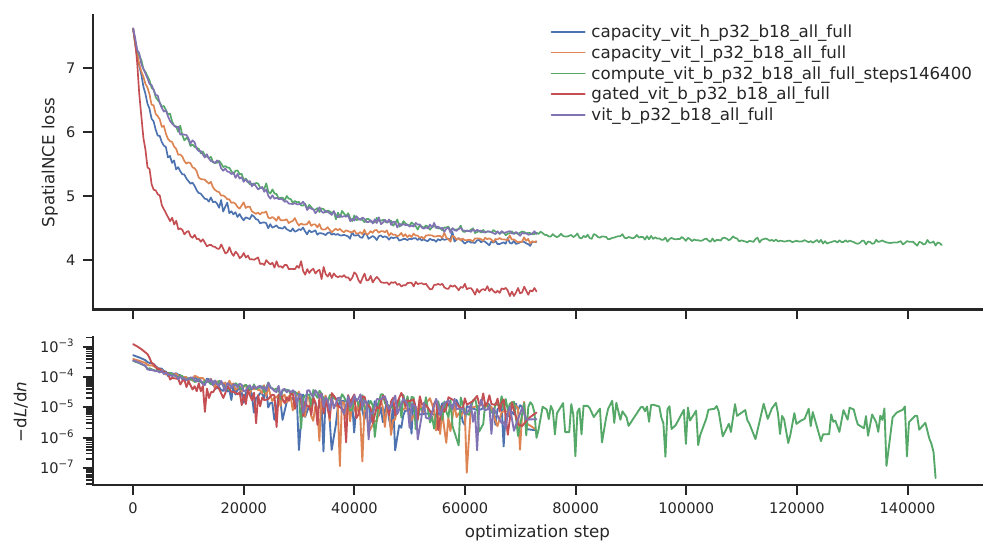}
	\end{center}
	\caption{\textbf{Training trajectories for five $B=18$ full-coverage configurations from \textbf{E5--E7} and their shared baseline.}
		The top panel shows SpatialNCE loss, and the bottom panel shows the gradient of the smoothed loss trajectory.
		The loss continues to decrease at the doubled budget of \num{146400} steps.
		See \cref{tab:s_inventory} for the corresponding configurations.
	}
	\label{fig:s_training}
\end{figure}

\Cref{fig:s_training} shows standard and doubled-compute ViT-B, capacity ViT-L/H, and subject-gated ViT-B trajectories.
Loss trajectories were smoothed with a Savitzky--Golay filter using a window of \num{1000} steps before computing their gradients.
SpatialNCE loss continues to decrease at the doubled budget of \num{146400} steps, indicating that training has not reached a loss plateau within the tested compute range.
Higher-capacity models reach lower SpatialNCE loss values over the same optimization budget.

\subsection{Alternative held-out-subject assignments}\label{app:altholdout}

\begin{figure}[h]
	\begin{center}
		\includegraphics[width=0.5\textwidth]{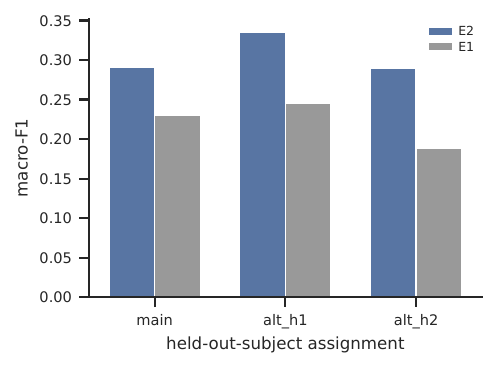}
	\end{center}
	\caption{
		\textbf{Macro-F1 for the $B=18$ configurations of \textbf{E1} and \textbf{E2} under two alternative unseen-subject assignments.}
		Absolute performance varies across assignments, while the direction and approximate magnitude of the difference between \textbf{E1} and \textbf{E2} are preserved.
	}
	\label{fig:s_altholdout}
\end{figure}

\Cref{fig:s_altholdout} compares the $B=18$ configurations of \textbf{E1} and \textbf{E2} under the primary unseen-subject assignment and two alternative assignments.
The two alternative subject triples are mutually disjoint, disjoint from the primary triple, and restricted to subjects with a coronal cutting plane.
Scores are computed over the 188 classes represented in all three assignments to make the results directly comparable across assignments.
This common class set causes the values in \cref{fig:s_altholdout} to differ slightly from the corresponding scores reported in the main text.
Re-evaluating all eligible \textbf{E1} configurations on an alternative unseen-subject triple also left the subject-count effect unresolved, with a ViT-B slope of $\beta_B=0.0131\,[-0.0164,0.0429]$ logit macro-F1 per doubling of $B$.

\subsection{Reference encoders}\label{app:reference}

\begin{table}[t]
	\caption{\textbf{Reference encoders evaluated as frozen feature extractors using the same probe and unseen-subject evaluation protocol.}
		Our rows correspond to the $B=18$ configurations of \textbf{E6} at each model capacity.
		The comparison provides an absolute performance reference across models with different architectures, pretraining data, and computational budgets.
		The chance row states where the macro-F1 scale begins under the same per-subject class denominator.
	}
	\label{tab:reference}
	\begin{center}
		\begin{tabular}{@{}llr@{}}
			\toprule
			\textbf{Encoder}     & \textbf{Reference}     & \textbf{Macro-F1} \\
			\midrule
			Uniform random guess & ---                    & $0.004$           \\
			\midrule
			DINOv2 (ViT-B/14)    & \citet{Oquab2024}      & $0.168$           \\
			GigaPath (ViT-G/16)  & \citet{Xu2024}         & $0.220$           \\
			UNI (ViT-L/16)       & \citet{Chen2024}       & $0.225$           \\
			Virchow2 (ViT-H/14)  & \citet{Zimmermann2024} & $0.251$           \\
			\midrule
			ViT-B/32             & ours                   & $0.300$           \\
			ViT-L/32             & ours                   & $0.311$           \\
			ViT-H/32             & ours                   & $0.316$           \\
			\bottomrule
		\end{tabular}
	\end{center}
\end{table}

\Cref{tab:reference} compares published vision and pathology encoders with the $B=18$ configurations of \textbf{E6} under the same frozen-feature probe and unseen-subject evaluation protocol.
Each $2048\times2048$-pixel input was split into sixteen $512\times512$ tiles, converted from grayscale to RGB, and processed natively (224 pixels for UNI/Virchow2, 518 for DINOv2, and resize to 256 then center-crop to 224 for GigaPath); embeddings were averaged.
Differences in architecture, pretraining, physical resolution, and modified-silver versus H\&E staining make this an absolute-performance reference only.
Among the evaluated reference encoders, Virchow2 achieves the highest macro-F1 at $0.251$, compared with $0.300$, $0.311$, and $0.316$ for our ViT-B/32, ViT-L/32, and ViT-H/32 models, respectively.
A uniform predictor over the 191-class vocabulary yields $0.004$ macro-F1 averaged across the three unseen subjects.

\section{Remaining limitations}\label{app:limitations}

The study has several limitations.
The subject pool is finite, and subject subsets at different values of $B$ overlap, so comparisons across subject-count conditions are not based on fully independent sets of subjects.
The primary evaluation uses three unseen subjects, although the main comparisons retain their direction under two additional held-out-subject assignments (\cref{app:altholdout}).
Each configuration was pretrained with a single optimization seed because of the computational cost of the study.
Replication across independently composed subject subsets captures sensitivity to subject selection but does not separately quantify optimization-seed variability.
The coverage manipulation in \textbf{E4} changes both sampled anatomy and the spatial distribution of pairwise relationships entering SpatialNCE.
Subject gating in \textbf{E7} directly changes the composition and total positive weight of the SpatialNCE objective.
The corresponding results therefore do not isolate all contributing mechanisms individually, as discussed in \cref{sec:r4}.
\textbf{E1} evaluates fixed-sample subject scaling at a single budget of approximately \num{450000} unique locations and over $B=1$--$18$.
Subject diversity may have a different effect at substantially smaller sample budgets or when each subject is sampled more sparsely.
All downstream results use linear probes on frozen representations and cytoarchitectonic area labels.
More expressive readouts may recover representation structure that is not linearly accessible, although the familiarity effect in \cref{sec:r1} shows that the linear probe is sensitive to substantial inter-subject differences.

\section{Complete run inventory}\label{app:inventory_table}

\Cref{tab:s_inventory} lists all 93 pretraining runs conducted in this study.
A pretraining run denotes one optimization trajectory, whereas a configuration in the statistical analysis denotes one retained checkpoint evaluated on the common primary unseen subjects.
Four runs from \textbf{E8} use alternative held-out-subject assignments and are excluded from the common-subject full-grid fit.
Each of the seven doubled-budget runs in \textbf{E5} contributes an additional retained half-budget checkpoint without requiring a separate optimization trajectory.
The full-grid fit therefore contains $93-4+7=96$ configurations and $96\times3=288$ subject-level scores.
The fixed-sample fit for \textbf{E1} contains 18 configurations: nine ViT-B, six ViT-H, and three ViT-L, yielding 54 subject-level scores.

\setlength{\LTcapwidth}{\textwidth}
\begingroup\setlength{\tabcolsep}{2pt}\renewcommand{\arraystretch}{0.8}
\begin{longtable}{@{}>{\scriptsize}l>{\scriptsize}l>{\scriptsize}c>{\scriptsize}r>{\scriptsize}l>{\scriptsize}r>{\scriptsize}r>{\scriptsize}c>{\scriptsize}c@{}}
	\caption{\textbf{Inventory of all 93 pretraining runs, grouped by experiment.} \emph{Steps} gives the completed optimization budget. The seven runs at \num{146400} steps also contribute a retained checkpoint at half that budget. \emph{Gated} marks the runs whose objective admits no cross-subject positives. \emph{Fit} marks inclusion in the full-grid model. The four runs on alternative held-out triples are analyzed only in \cref{app:altholdout}.}\label{tab:s_inventory} \\
	\toprule
	\textbf{exp.} & \textbf{run}                                                             & \textbf{arch} & \textbf{$B$} & \textbf{sampling} & \textbf{$N_{\mathrm{unique}}$} & \textbf{steps} & \textbf{gated} & \textbf{fit} \\
	\midrule
	\endfirsthead
	\toprule
	\textbf{exp.} & \textbf{run}                                                             & \textbf{arch} & \textbf{$B$} & \textbf{sampling} & \textbf{$N_{\mathrm{unique}}$} & \textbf{steps} & \textbf{gated} & \textbf{fit} \\
	\midrule
	\endhead
	\midrule
	\multicolumn{9}{r@{}}{\scriptsize\emph{continued on next page}} \\
	\endfoot
	\bottomrule
	\endlastfoot
	E1            & \texttt{vit\_b\_p32\_b01\_subset03\_iso\_unique}                         & B             & 1            & fixed budget      & \num{449924}                   & \num{73200}    & \ding{55}      & \ding{51}    \\
	E1            & \texttt{vit\_b\_p32\_b01\_subset04\_iso\_unique}                         & B             & 1            & fixed budget      & \num{450792}                   & \num{73200}    & \ding{55}      & \ding{51}    \\
	E1            & \texttt{vit\_b\_p32\_b02\_subset01\_iso\_unique}                         & B             & 2            & fixed budget      & \num{449883}                   & \num{73200}    & \ding{55}      & \ding{51}    \\
	E1            & \texttt{vit\_b\_p32\_b02\_subset02\_iso\_unique}                         & B             & 2            & fixed budget      & \num{450825}                   & \num{73200}    & \ding{55}      & \ding{51}    \\
	E1            & \texttt{vit\_b\_p32\_b04\_subset01\_iso\_unique}                         & B             & 4            & fixed budget      & \num{450404}                   & \num{73200}    & \ding{55}      & \ding{51}    \\
	E1            & \texttt{vit\_b\_p32\_b04\_subset02\_iso\_unique}                         & B             & 4            & fixed budget      & \num{449245}                   & \num{73200}    & \ding{55}      & \ding{51}    \\
	E1            & \texttt{vit\_b\_p32\_b08\_subset01\_iso\_unique}                         & B             & 8            & fixed budget      & \num{448861}                   & \num{73200}    & \ding{55}      & \ding{51}    \\
	E1            & \texttt{vit\_b\_p32\_b08\_subset02\_iso\_unique}                         & B             & 8            & fixed budget      & \num{450661}                   & \num{73200}    & \ding{55}      & \ding{51}    \\
	E1            & \texttt{vit\_b\_p32\_b18\_all\_iso\_unique}                              & B             & 18           & fixed budget      & \num{449794}                   & \num{73200}    & \ding{55}      & \ding{51}    \\
	E2            & \texttt{vit\_b\_p32\_b01\_subset01\_alt10}                               & B             & 1            & every 10th        & \num{52461}                    & \num{73200}    & \ding{55}      & \ding{51}    \\
	E2            & \texttt{vit\_b\_p32\_b01\_subset02\_alt10}                               & B             & 1            & every 10th        & \num{64795}                    & \num{73200}    & \ding{55}      & \ding{51}    \\
	E2            & \texttt{vit\_b\_p32\_b01\_subset03\_alt10}                               & B             & 1            & every 10th        & \num{48732}                    & \num{73200}    & \ding{55}      & \ding{51}    \\
	E2            & \texttt{vit\_b\_p32\_b01\_subset04\_alt10}                               & B             & 1            & every 10th        & \num{43829}                    & \num{73200}    & \ding{55}      & \ding{51}    \\
	E2            & \texttt{vit\_b\_p32\_b01\_subset01\_full}                                & B             & 1            & all               & \num{532668}                   & \num{73200}    & \ding{55}      & \ding{51}    \\
	E2            & \texttt{vit\_b\_p32\_b01\_subset02\_full}                                & B             & 1            & all               & \num{668055}                   & \num{73200}    & \ding{55}      & \ding{51}    \\
	E2            & \texttt{vit\_b\_p32\_b01\_subset03\_full}                                & B             & 1            & all               & \num{500528}                   & \num{73200}    & \ding{55}      & \ding{51}    \\
	E2            & \texttt{vit\_b\_p32\_b01\_subset04\_full}                                & B             & 1            & all               & \num{461204}                   & \num{73200}    & \ding{55}      & \ding{51}    \\
	E2            & \texttt{vit\_b\_p32\_b02\_subset01\_alt10}                               & B             & 2            & every 10th        & \num{107329}                   & \num{73200}    & \ding{55}      & \ding{51}    \\
	E2            & \texttt{vit\_b\_p32\_b02\_subset02\_alt10}                               & B             & 2            & every 10th        & \num{115616}                   & \num{73200}    & \ding{55}      & \ding{51}    \\
	E2            & \texttt{vit\_b\_p32\_b02\_subset03\_alt10}                               & B             & 2            & every 10th        & \num{101308}                   & \num{73200}    & \ding{55}      & \ding{51}    \\
	E2            & \texttt{vit\_b\_p32\_b02\_subset04\_alt10}                               & B             & 2            & every 10th        & \num{111274}                   & \num{73200}    & \ding{55}      & \ding{51}    \\
	E2            & \texttt{vit\_b\_p32\_b02\_subset01\_full}                                & B             & 2            & all               & \num{1090235}                  & \num{73200}    & \ding{55}      & \ding{51}    \\
	E2            & \texttt{vit\_b\_p32\_b02\_subset02\_full}                                & B             & 2            & all               & \num{1177529}                  & \num{73200}    & \ding{55}      & \ding{51}    \\
	E2            & \texttt{vit\_b\_p32\_b02\_subset03\_full}                                & B             & 2            & all               & \num{1034343}                  & \num{73200}    & \ding{55}      & \ding{51}    \\
	E2            & \texttt{vit\_b\_p32\_b02\_subset04\_full}                                & B             & 2            & all               & \num{1133526}                  & \num{73200}    & \ding{55}      & \ding{51}    \\
	E2            & \texttt{vit\_b\_p32\_b04\_subset01\_alt10}                               & B             & 4            & every 10th        & \num{218659}                   & \num{73200}    & \ding{55}      & \ding{51}    \\
	E2            & \texttt{vit\_b\_p32\_b04\_subset02\_alt10}                               & B             & 4            & every 10th        & \num{214133}                   & \num{73200}    & \ding{55}      & \ding{51}    \\
	E2            & \texttt{vit\_b\_p32\_b04\_subset03\_alt10}                               & B             & 4            & every 10th        & \num{212405}                   & \num{73200}    & \ding{55}      & \ding{51}    \\
	E2            & \texttt{vit\_b\_p32\_b04\_subset04\_alt10}                               & B             & 4            & every 10th        & \num{231730}                   & \num{73200}    & \ding{55}      & \ding{51}    \\
	E2            & \texttt{vit\_b\_p32\_b04\_subset01\_full}                                & B             & 4            & all               & \num{2233423}                  & \num{73200}    & \ding{55}      & \ding{51}    \\
	E2            & \texttt{vit\_b\_p32\_b04\_subset02\_full}                                & B             & 4            & all               & \num{2188587}                  & \num{73200}    & \ding{55}      & \ding{51}    \\
	E2            & \texttt{vit\_b\_p32\_b04\_subset03\_full}                                & B             & 4            & all               & \num{2177018}                  & \num{73200}    & \ding{55}      & \ding{51}    \\
	E2            & \texttt{vit\_b\_p32\_b04\_subset04\_full}                                & B             & 4            & all               & \num{2355517}                  & \num{73200}    & \ding{55}      & \ding{51}    \\
	E2            & \texttt{vit\_b\_p32\_b08\_subset01\_alt10}                               & B             & 8            & every 10th        & \num{425687}                   & \num{73200}    & \ding{55}      & \ding{51}    \\
	E2            & \texttt{vit\_b\_p32\_b08\_subset02\_alt10}                               & B             & 8            & every 10th        & \num{451662}                   & \num{73200}    & \ding{55}      & \ding{51}    \\
	E2            & \texttt{vit\_b\_p32\_b08\_subset01\_full}                                & B             & 8            & all               & \num{4356763}                  & \num{73200}    & \ding{55}      & \ding{51}    \\
	E2            & \texttt{vit\_b\_p32\_b08\_subset02\_full}                                & B             & 8            & all               & \num{4605682}                  & \num{73200}    & \ding{55}      & \ding{51}    \\
	E2            & \texttt{vit\_b\_p32\_b18\_all\_alt10}                                    & B             & 18           & every 10th        & \num{962026}                   & \num{73200}    & \ding{55}      & \ding{51}    \\
	E2            & \texttt{vit\_b\_p32\_b18\_all\_full}                                     & B             & 18           & all               & \num{9834115}                  & \num{73200}    & \ding{55}      & \ding{51}    \\
	E3            & \texttt{vit\_b\_p32\_b01\_subset03\_alt33}                               & B             & 1            & every 3rd         & \num{164378}                   & \num{73200}    & \ding{55}      & \ding{51}    \\
	E3            & \texttt{vit\_b\_p32\_b09\_subset01\_alt10}                               & B             & 9            & every 10th        & \num{472504}                   & \num{73200}    & \ding{55}      & \ding{51}    \\
	E3            & \texttt{vit\_b\_p32\_b09\_subset01\_alt33}                               & B             & 9            & every 3rd         & \num{1579631}                  & \num{73200}    & \ding{55}      & \ding{51}    \\
	E3            & \texttt{vit\_b\_p32\_b09\_subset01\_full}                                & B             & 9            & all               & \num{4816705}                  & \num{73200}    & \ding{55}      & \ding{51}    \\
	E3            & \texttt{vit\_b\_p32\_b18\_all\_alt33}                                    & B             & 18           & every 3rd         & \num{3225383}                  & \num{73200}    & \ding{55}      & \ding{51}    \\
	\pagebreak
	E4            & \texttt{vit\_b\_p32\_b01\_coronal\_subset01\_alt10}                      & B             & 1            & every 10th        & \num{48732}                    & \num{73200}    & \ding{55}      & \ding{51}    \\
	E4            & \texttt{vit\_b\_p32\_b01\_coronal\_subset01\_fragmented04}               & B             & 1            & 4 blocks          & \num{49036}                    & \num{73200}    & \ding{55}      & \ding{51}    \\
	E4            & \texttt{vit\_b\_p32\_b01\_coronal\_subset01\_full}                       & B             & 1            & all               & \num{500528}                   & \num{73200}    & \ding{55}      & \ding{51}    \\
	E4            & \texttt{vit\_b\_p32\_b01\_coronal\_subset01\_localized\_matched\_end}    & B             & 1            & loc.\ anterior    & \num{48872}                    & \num{73200}    & \ding{55}      & \ding{51}    \\
	E4            & \texttt{vit\_b\_p32\_b01\_coronal\_subset01\_localized\_matched\_middle} & B             & 1            & loc.\ center      & \num{50094}                    & \num{73200}    & \ding{55}      & \ding{51}    \\
	E4            & \texttt{vit\_b\_p32\_b01\_coronal\_subset01\_localized\_matched\_start}  & B             & 1            & loc.\ posterior   & \num{49110}                    & \num{73200}    & \ding{55}      & \ding{51}    \\
	E4            & \texttt{vit\_b\_p32\_b04\_coronal\_subset01\_alt10}                      & B             & 4            & every 10th        & \num{219403}                   & \num{73200}    & \ding{55}      & \ding{51}    \\
	E4            & \texttt{vit\_b\_p32\_b04\_coronal\_subset01\_fragmented04}               & B             & 4            & 4 blocks          & \num{223273}                   & \num{73200}    & \ding{55}      & \ding{51}    \\
	E4            & \texttt{vit\_b\_p32\_b04\_coronal\_subset01\_full}                       & B             & 4            & all               & \num{2237676}                  & \num{73200}    & \ding{55}      & \ding{51}    \\
	E4            & \texttt{vit\_b\_p32\_b04\_coronal\_subset01\_localized\_matched\_end}    & B             & 4            & loc.\ anterior    & \num{221468}                   & \num{73200}    & \ding{55}      & \ding{51}    \\
	E4            & \texttt{vit\_b\_p32\_b04\_coronal\_subset01\_localized\_matched\_middle} & B             & 4            & loc.\ center      & \num{223022}                   & \num{73200}    & \ding{55}      & \ding{51}    \\
	E4            & \texttt{vit\_b\_p32\_b04\_coronal\_subset01\_localized\_matched\_start}  & B             & 4            & loc.\ posterior   & \num{221797}                   & \num{73200}    & \ding{55}      & \ding{51}    \\
	E4            & \texttt{vit\_b\_p32\_b13\_coronal\_all\_alt10}                           & B             & 13           & every 10th        & \num{701132}                   & \num{73200}    & \ding{55}      & \ding{51}    \\
	E4            & \texttt{vit\_b\_p32\_b13\_coronal\_all\_fragmented04}                    & B             & 13           & 4 blocks          & \num{706946}                   & \num{73200}    & \ding{55}      & \ding{51}    \\
	E4            & \texttt{vit\_b\_p32\_b13\_coronal\_all\_fragmented16}                    & B             & 13           & 16 blocks         & \num{711367}                   & \num{73200}    & \ding{55}      & \ding{51}    \\
	E4            & \texttt{vit\_b\_p32\_b13\_coronal\_all\_full}                            & B             & 13           & all               & \num{7151524}                  & \num{73200}    & \ding{55}      & \ding{51}    \\
	E4            & \texttt{vit\_b\_p32\_b13\_coronal\_all\_localized\_matched\_end}         & B             & 13           & loc.\ anterior    & \num{708859}                   & \num{73200}    & \ding{55}      & \ding{51}    \\
	E4            & \texttt{vit\_b\_p32\_b13\_coronal\_all\_localized\_matched\_middle}      & B             & 13           & loc.\ center      & \num{712904}                   & \num{73200}    & \ding{55}      & \ding{51}    \\
	E4            & \texttt{vit\_b\_p32\_b13\_coronal\_all\_localized\_matched\_start}       & B             & 13           & loc.\ posterior   & \num{707964}                   & \num{73200}    & \ding{55}      & \ding{51}    \\
	E5            & \texttt{compute\_vit\_b\_p32\_b01\_subset03\_alt10\_steps146400}         & B             & 1            & every 10th        & \num{48732}                    & \num{146400}   & \ding{55}      & \ding{51}    \\
	E5            & \texttt{compute\_vit\_b\_p32\_b01\_subset03\_full\_steps146400}          & B             & 1            & all               & \num{500528}                   & \num{146400}   & \ding{55}      & \ding{51}    \\
	E5            & \texttt{compute\_vit\_b\_p32\_b04\_subset01\_alt10\_steps146400}         & B             & 4            & every 10th        & \num{218659}                   & \num{146400}   & \ding{55}      & \ding{51}    \\
	E5            & \texttt{compute\_vit\_b\_p32\_b04\_subset01\_full\_steps146400}          & B             & 4            & all               & \num{2233423}                  & \num{146400}   & \ding{55}      & \ding{51}    \\
	E5            & \texttt{compute\_vit\_b\_p32\_b18\_all\_alt10\_steps146400}              & B             & 18           & every 10th        & \num{962026}                   & \num{146400}   & \ding{55}      & \ding{51}    \\
	E5            & \texttt{compute\_vit\_b\_p32\_b18\_all\_alt33\_steps146400}              & B             & 18           & every 3rd         & \num{3225383}                  & \num{146400}   & \ding{55}      & \ding{51}    \\
	E5            & \texttt{compute\_vit\_b\_p32\_b18\_all\_full\_steps146400}               & B             & 18           & all               & \num{9834115}                  & \num{146400}   & \ding{55}      & \ding{51}    \\
	E6            & \texttt{capacity\_vit\_h\_p32\_b01\_subset03\_alt10}                     & H             & 1            & every 10th        & \num{48732}                    & \num{73200}    & \ding{55}      & \ding{51}    \\
	E6            & \texttt{capacity\_vit\_h\_p32\_b01\_subset03\_full}                      & H             & 1            & all               & \num{500528}                   & \num{73200}    & \ding{55}      & \ding{51}    \\
	E6            & \texttt{capacity\_vit\_h\_p32\_b01\_subset03\_iso\_unique}               & H             & 1            & fixed budget      & \num{449924}                   & \num{73200}    & \ding{55}      & \ding{51}    \\
	E6            & \texttt{capacity\_vit\_h\_p32\_b01\_subset04\_iso\_unique}               & H             & 1            & fixed budget      & \num{450792}                   & \num{73200}    & \ding{55}      & \ding{51}    \\
	E6            & \texttt{capacity\_vit\_l\_p32\_b01\_subset03\_iso\_unique}               & L             & 1            & fixed budget      & \num{449924}                   & \num{73200}    & \ding{55}      & \ding{51}    \\
	E6            & \texttt{capacity\_vit\_h\_p32\_b02\_subset01\_iso\_unique}               & H             & 2            & fixed budget      & \num{449883}                   & \num{73200}    & \ding{55}      & \ding{51}    \\
	E6            & \texttt{capacity\_vit\_h\_p32\_b04\_subset01\_iso\_unique}               & H             & 4            & fixed budget      & \num{450404}                   & \num{73200}    & \ding{55}      & \ding{51}    \\
	E6            & \texttt{capacity\_vit\_l\_p32\_b04\_subset01\_iso\_unique}               & L             & 4            & fixed budget      & \num{450404}                   & \num{73200}    & \ding{55}      & \ding{51}    \\
	E6            & \texttt{capacity\_vit\_h\_p32\_b08\_subset01\_iso\_unique}               & H             & 8            & fixed budget      & \num{448861}                   & \num{73200}    & \ding{55}      & \ding{51}    \\
	E6            & \texttt{capacity\_vit\_h\_p32\_b18\_all\_alt10}                          & H             & 18           & every 10th        & \num{962026}                   & \num{73200}    & \ding{55}      & \ding{51}    \\
	E6            & \texttt{capacity\_vit\_h\_p32\_b18\_all\_full}                           & H             & 18           & all               & \num{9834115}                  & \num{73200}    & \ding{55}      & \ding{51}    \\
	E6            & \texttt{capacity\_vit\_l\_p32\_b18\_all\_full}                           & L             & 18           & all               & \num{9834115}                  & \num{73200}    & \ding{55}      & \ding{51}    \\
	E6            & \texttt{capacity\_vit\_h\_p32\_b18\_all\_iso\_unique}                    & H             & 18           & fixed budget      & \num{449794}                   & \num{73200}    & \ding{55}      & \ding{51}    \\
	E6            & \texttt{capacity\_vit\_l\_p32\_b18\_all\_iso\_unique}                    & L             & 18           & fixed budget      & \num{449794}                   & \num{73200}    & \ding{55}      & \ding{51}    \\
	E7            & \texttt{gated\_vit\_b\_p32\_b02\_subset01\_full}                         & B             & 2            & all               & \num{1090235}                  & \num{73200}    & \ding{51}      & \ding{51}    \\
	E7            & \texttt{gated\_vit\_b\_p32\_b04\_subset01\_full}                         & B             & 4            & all               & \num{2233423}                  & \num{73200}    & \ding{51}      & \ding{51}    \\
	E7            & \texttt{gated\_vit\_b\_p32\_b08\_subset01\_full}                         & B             & 8            & all               & \num{4356763}                  & \num{73200}    & \ding{51}      & \ding{51}    \\
	E7            & \texttt{gated\_vit\_b\_p32\_b18\_all\_alt10}                             & B             & 18           & every 10th        & \num{962026}                   & \num{73200}    & \ding{51}      & \ding{51}    \\
	E7            & \texttt{gated\_vit\_b\_p32\_b18\_all\_full}                              & B             & 18           & all               & \num{9834115}                  & \num{73200}    & \ding{51}      & \ding{51}    \\
	E8            & \texttt{vit\_b\_p32\_b18\_alth1\_all\_full}                              & B             & 18           & all               & \num{10041128}                 & \num{73200}    & \ding{55}      & \ding{55}    \\
	E8            & \texttt{vit\_b\_p32\_b18\_alth2\_all\_full}                              & B             & 18           & all               & \num{9951511}                  & \num{73200}    & \ding{55}      & \ding{55}    \\
	E8            & \texttt{vit\_b\_p32\_b18\_alth1\_all\_iso\_unique}                       & B             & 18           & fixed budget      & \num{448344}                   & \num{73200}    & \ding{55}      & \ding{55}    \\
	E8            & \texttt{vit\_b\_p32\_b18\_alth2\_all\_iso\_unique}                       & B             & 18           & fixed budget      & \num{449438}                   & \num{73200}    & \ding{55}      & \ding{55}    \\
\end{longtable}
\endgroup

\end{document}